\pdfoutput=1
\documentclass[11pt]{article}

\usepackage{styles/acl}

\usepackage{times}
\usepackage{latexsym}
\usepackage[T1]{fontenc}
\usepackage[utf8]{inputenc}
\usepackage{microtype}
\usepackage{inconsolata}
\usepackage{graphicx}

\usepackage{caption}
\usepackage{subcaption}

\usepackage{graphicx}
\usepackage{booktabs}   % for \toprule, \midrule, and \bottomrule
\usepackage{array}
\usepackage{verbatim}   % for verbatim-like environments
\usepackage{adjustbox}

\usepackage{amssymb}
\usepackage{pifont}

\newcommand{\reffig}[1]{Figure~\ref{#1}}
\newcommand{\reftbl}[1]{Table~\ref{#1}}
\newcommand{\refsec}[1]{Section~\ref{#1}}

\newcommand{\xmark}{\ding{55}}

\newcommand{\model}{GeoCoder}
\newcommand{\rag}{RAG-GeoCoder}

\title{\model: Solving Geometry Problems by Generating Modular Code through Vision-Language Models}

\author{
 \textbf{Aditya Sharma\textsuperscript{1,2}},
 \textbf{Aman Dalmia\textsuperscript{1,3}},
 \textbf{Mehran Kazemi\textsuperscript{5}},
 \textbf{Amal Zouaq\textsuperscript{1,2}},
 \textbf{Christopher J. Pal\textsuperscript{1,2,4}}
\\
\\
 \textsuperscript{1}Mila,
 \textsuperscript{2}Polytechnique Montréal,
 \textsuperscript{3}Université de Montréal,
 \textsuperscript{4}Canada CIFAR AI Chair,
 \textsuperscript{5}Google Deepmind
\\
 \small{
   \textbf{Correspondence:} \href{mailto:aditya.sharma@mila.quebec}{aditya.sharma@mila.quebec}
 }
}

\begin{document}
\maketitle

\begin{abstract}
% While large language models (LLMs) excel in text-based mathematical reasoning tasks, many mathematical problems incorporate both textual and visual components.

Geometry problem-solving demands advanced reasoning abilities to process multimodal inputs and employ mathematical knowledge effectively.
Vision-language models (VLMs) have made significant progress in various multimodal tasks. 
Yet, they still struggle with geometry problems and are significantly limited by their inability to perform mathematical operations not seen during pre-training, such as calculating the 
% perimeter of an object
cosine of an arbitrary angle, and by difficulties in correctly applying relevant geometry formulas.
To overcome these challenges, we present \model, which leverages modular code-finetuning to generate and execute code using a predefined geometry function library. By executing the code, we achieve accurate and deterministic calculations, contrasting the stochastic nature of autoregressive token prediction, while the function library minimizes errors in formula usage.
% To address these limitations, we introduce \model\ which employs a two-step training approach: (1) first, creating a high-quality `code' instruction-tuning dataset by few-shot prompting a teacher LLM (large language model) to generate modular code that uses a predefined geometric function library, and (2) then using this data to fine-tune a smaller student VLM for modular code generation in complex geometry tasks.
We also propose a multimodal retrieval-augmented variant of \model, named \rag, which incorporates a non-parametric memory module for retrieving functions from the geometry library, thereby reducing reliance on parametric memory.
% This enables \rag\ to handle more extensive libraries and accommodate library updates, while consistently performing better than \model.
% Furthermore, we propose a multi-modal retrieval-augmented version of \model, called \rag, which can adapt to changes in the function library without needing additional fine-tuning.
% Our proposed modular code-finetuning improves the geometric reasoning performance of VLMs over other fine-tuning approaches and adds interpretability through templated print statements.
Our modular code-finetuning approach enhances the geometric reasoning capabilities of VLMs, yielding an average improvement of over $16\%$ across various question complexities on the GeomVerse dataset compared to other fine-tuning methods.
\end{abstract}
\section{Introduction}
\label{sec:introduction}

% With the increased adoption of large language models (LLMs),% in educational domains, 
% mathematical reasoning represents a critical challenge and remains a prominent aim for artificial intelligence research.  
% Recent LLMs have shown remarkable performance on textual mathematical problems \citep{llama31,gpt-4o}. 
% This study centers on the geometry question answering and reasoning task, which requires interpreting a given figure, comprehending the question, and applying the appropriate geometry formulas to derive the answer. 

Geometry problem solving requires interpreting a figure, comprehending a question, and applying the appropriate geometry formulas to derive the answer.
This presents an ideal test-bed for evaluating multimodal models' visual and mathematical reasoning capabilities, with practical relevance in fields such as education, architecture, and engineering.
% This tests the advanced reasoning abilities of multimodal language models to process multimodal inputs and employ mathematical knowledge effectively.

\begin{figure}[h]
\centering
\includegraphics[width=0.35\textwidth]{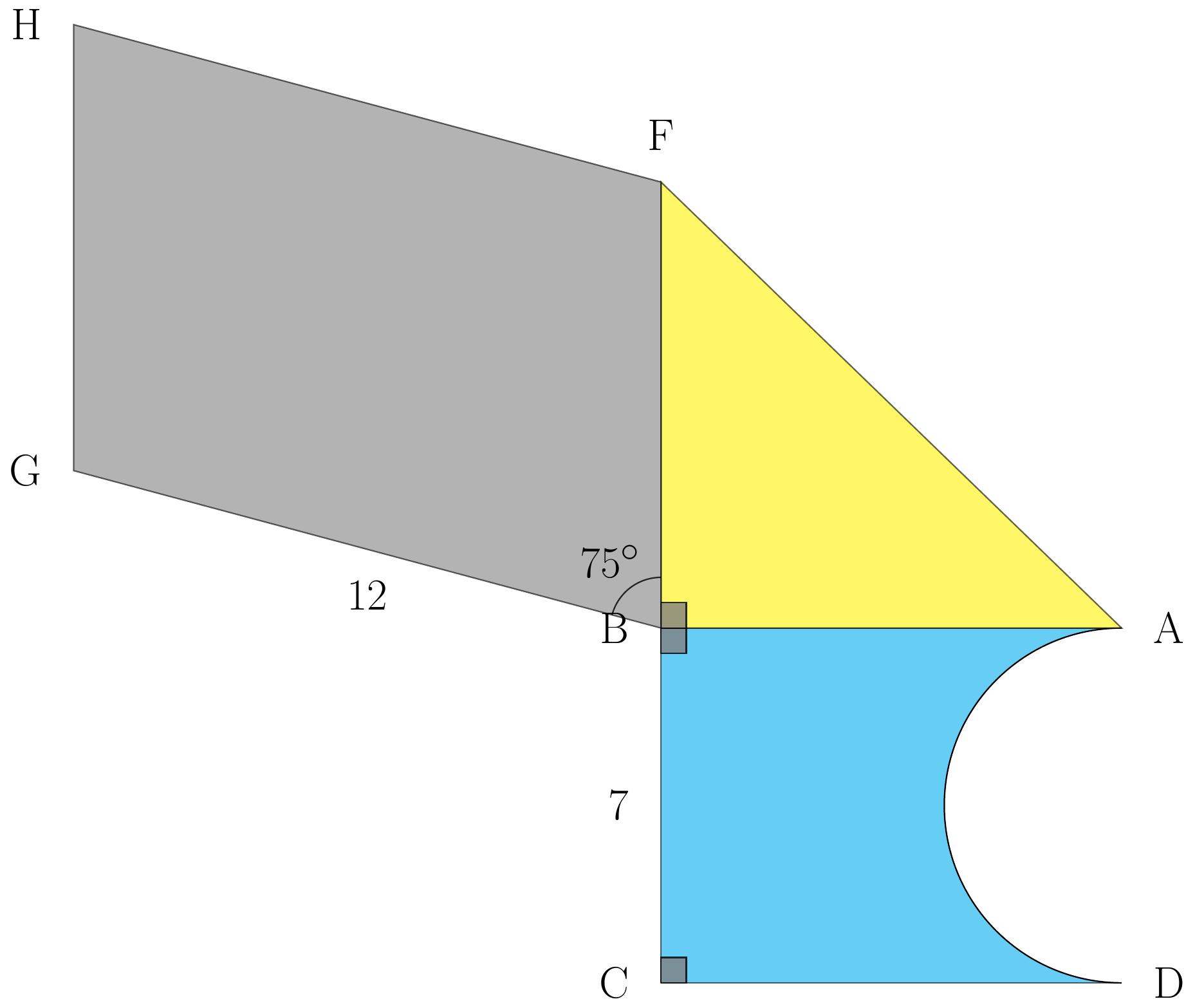}
\caption{Sample geometry problem from the GeomVerse \citep{geomverse} dataset. \textbf{Question}: If the ABCD shape is a rectangle where a semi-circle has been removed from one side of it, the area of the BAF right triangle is 40 and the area of the BGHF parallelogram is 102, compute the area of the ABCD shape.}
\label{fig:dataset}
\end{figure}

Various vision-language models (VLMs) like GPT-4V \citep{gpt4}, GPT-4o \citep{gpt-4o}, Gemini \citep{gemini}, PaLI \citep{pali} and LLaVA \citep{llava} have been introduced and benchmarked on tasks involving geometric reasoning \citep{gllava,geomverse}.
Despite this progress, the inherent nature of auto-regressive language model training, which focuses on next-token prediction, presents challenges in modelling mathematical reasoning that requires precise calculations and the correct application of formulas \citep{geomverse, mario}.
While language models can handle computations they were exposed to during training, such as cos 45, they tend to struggle with unseen calculations, for instance, cos 18.
The Chain-of-Thought (CoT) approach aims to improve reasoning in language models by eliciting a step-by-step natural language reasoning along with the final answer \citep{chain_of_thought} but suffers from the same problems of imprecise computations and wrong formula usage \citep{tora,geomverse}.
This problem is especially aggravated when the geometry problems require multi-hop mathematical reasoning over figures with multiple complex shapes like in the GeomVerse dataset \citep{geomverse}, as seen in figure \ref{fig:dataset}.

Mathematical code generation, where a model generates code that is then executed to obtain the solution to a mathematical problem, addresses the issue of inaccurate calculations \citep{pot, mathcoder, mario}. Code execution ensures precise and deterministic results, unlike the stochastic behaviour of autoregressive next-token prediction. However, this method does not fully resolve the issue of incorrect formula application, as the language model relies on its parametric memory to recall and implement the relevant formulas.

In this study, we answer the following research question: 
% \textit{"Can VLMs achieve improved multimodal mathematical reasoning by employing modular code generation and execution, rather than relying on auto-regressive answer generation from parametric memory?"}
\textit{"Can VLMs enhance their multimodal mathematical reasoning capabilities by using modular code generation and execution, instead of relying on autoregressive answer generation?"}
We propose \model, a multi-modal modular code generation approach for geometric reasoning using VLMs, which generates code that includes calls to predefined mathematical functions and mitigates this limitation by offloading the responsibility of recalling formulas to a function library. % that applies the correct formulas to accurately compute results. 
Modular code generation offers the additional advantage of improving interpretability by including templated print statements in the function outputs.
Solving most geometry questions involves applying multiple functions in sequence, which produces a series of templated explanations outlining each step of the problem-solving process.
% An added advantage of modular code generation is interpretability through templated print statements in the functions. % that explain how the formula was used on the inputs to get the final answer.

%%%%%%%%%%%%%%%%%%%%%%%%%%%%%%%%%

%%%%%%%%%%%%%%%%%%%%%%%%%%%%%%%%%

While recent LLMs (large language models) have demonstrated remarkable performance in Python code generation tasks \citep{code_survey}, VLMs still encounter difficulties, largely due to the lack of datasets pairing images and questions with code.
To address this gap, we capitalize on the code generation capabilities of LLMs to construct the first multimodal code instruction-tuning dataset for solving geometry problems, and subsequently finetune a VLM on this data to create \model.
% In addition, we introduce a retrieval-augmented version of our model, named \rag, in which a multimodal retriever selects a set of functions from the library that are relevant for answering the geometry question and incorporates them into the VLM's prompt. The VLM then learns to choose the most suitable function from this set to answer the question. 
% In addition, we introduce a retrieval-augmented version of our model, named \rag, which benefits from two main advantages over \model: (1) it can leverage non-parametric function libraries, and (2) it can allow the VLM to adapt to updates in the function library without requiring further finetuning.
In addition, we introduce a retrieval-augmented version of our model, named \rag, which leverages a non-parametric function memory and generates modular code which uses functions more often than \model\ (see \refsec{subsec:funtion_use}).
Given its reliance on non-parametric memory, \rag\ has the potential to adapt to both modifications in the function library and significantly larger libraries. However, we leave further investigation of this capability for future work.
% This strategy allows the model to adapt to updates in the function library without requiring further finetuning.

% Prior works like ViperGPT \citep{vipergpt} and VizProg \citep{visual_programming} have used code generation for visual tasks such as visual question answering (VQA), but they rely solely on LLMs that process only the question text for generating code and later incorporate image information during code execution.
% In contrast, our approach builds a VLM that integrates geometric image information alongside the question text to produce modular code for solving geometry problems.

% Furthermore, we examine the capacity of VLMs to utilize a predefined function library while generating code for executing geometric reasoning tasks.

This work makes the following contributions:
\begin{enumerate}
% \item We release a first-of-its-kind multimodal dataset for code instruction-tuning for solving geometry problems, with 35k geometry questions and images from GeomVerse, paired with corresponding code generated by our approach. We also...
    % \item We introduce a two-step method for training VLMs to generate modular code for solving geometry problems by utilizing a predefined function library. The VLM trained via this process is referred to as \model.
    \item Our work presents a modular code-finetuning framework for VLM-based geometry problem-solving utilizing a predefined function library. The VLM trained via this process is referred to as \model.
    \item We propose a retrieval-augmented variant of \model, called \rag, which includes a non-parametric memory to reduce the VLM's reliance on its parametric memory for recalling geometry functions, thus increasing function usage. 
    % It also enables the VLM to accommodate larger function libraries and adapt to library updates.
    \item \model\ and \rag\ outperform models trained with alternate finetuning strategies on the GeomVerse dataset by more than $16\%$ on average across question complexities and by $14.2\%$ on the GeoQA-NO dataset (see \refsec{subsec:eval_datasets}).
    \item \model\ and \rag\  add interpretability to VLM-based geometry problem-solving through templated print statements in the proposed geometry function library.
    \item We release a first-of-its-kind multimodal dataset for code instruction-tuning for solving geometry problems, with geometry questions and images paired with corresponding code generated by our approach. This dataset contains $35k$ data points from GeomVerse \citep{geomverse} and $48k$ from Geo170k \citep{gllava}.
    
\end{enumerate}

\section{Related Work}
\label{sec:related_work}

In this section, we discuss a literature review of existing approaches. %Section \ref{subsec:related_geometry} describes approaches to solving geometry problems using VLMs, while section \ref{subsec:related_math_llm} talks about recent models that use large language models (LLMs) for generating code to solve non-visual mathematical problems. In section \ref{subsec:related_code_vlm}, we discuss VLM-based code generation approaches to solve general visual question-answering (VQA) problems.

%------------------------------------------------------------

\subsection{VLMs for solving geometry problems}
\label{subsec:related_geometry}

% First, we describe approaches to solving geometry problems using VLMs. 
Recent advancements have seen the development of various general-purpose VLMs \citep{llava,gpt-4o,gemini,llava_onevision,infimm}, but they continue to underperform on geometry tasks.
% While we focus on the inability of VLMs to perform calculations and use correct formulas, 
\citet{gllava} and \citet{mavis} argue that VLMs struggle to comprehend basic geometric visual elements accurately and release two large geometry instruction-tuning datasets called Geo170k and MAVIS, respectively.
\citet{geomverse} release a new large synthetic dataset called GeomVerse for solving multi-hop geometry problems that require complex reasoning. 
They benchmark state-of-the-art VLMs on this dataset and show that these models are not as capable in subjects like geometry, which require long chains of reasoning. 
Our geometry code-tuning dataset is built using the Geo170k and GeomVerse datasets.
Additionally, AlphaGeometry \citep{alphageometry} introduces a neuro-symbolic system that solves geometry theorem-proving problems. In contrast to our model which generates modular code through a VLM, AlphaGeometry's generation combines a VLM with a rule-bound deduction engine. 

\subsection{Code generating LLMs for non-visual mathematical reasoning}
\label{subsec:related_math_llm}

% Secondly, we discuss recent models that use large language models (LLMs) for generating code to solve non-visual mathematical problems. 
\citet{mathcoder} and  \citet{mario} suggest a methodology for producing new datasets comprising textual math problems and their code-based solutions using large LLMs as teachers for smaller LLMs. 
We modify this knowledge-distillation approach to address multimodal mathematical reasoning by using a teacher LLM to generate our geometry code instruction-tuning dataset. (see \refsec{subsec:geo_data_gen}). Additionally, we extend this approach to modular code generation by incorporating a predefined geometry function library, ensuring the model applies precise mathematical formulas instead of relying on its parametric memory.
% This approach of generating code from scratch is in contrast to our idea of generating modular code using predefined functions which enables the model to use exact mathematical formulae instead of relying on its parametric memory.
% In contrast to our model which focuses on multi-modal mathematical reasoning using VLMs, \citet{tora} focus on text-only mathematical problems and propose agents that interweave natural language reasoning with program-based tool use.
\citet{tora} propose text-only mathematical reasoning agents that interweave natural language rationale generation with program-based tool use, this is in contrast to our multimodal approach where the VLM generates modular code which is executed to get the answer as well as the rationale through templated print statements.
% \noindent \textbf{MathCoder:} \citet{mathcoder} suggest a methodology for producing new datasets comprising math problems and their code-based solutions. They fine-tune LLaMa models \citep{llama2} on this generated data to build a family of models capable of generating code-based solutions for solving math problems. This approach of generating code from scratch is in contrast to our idea of generating modular code using predefined functions which enables the model to use exact mathematical formulae instead of relying on its parametric memory. 

% \noindent \textbf{ToRA:} \citet{tora} propose Tool-integrated Reasoning Agents that interweave natural language reasoning with program-based tool use. They use GPT-4 \citet{gpt4} to collect interactive tool-use trajectories for mathematical problems and fine-tune using these. This is similar to our knowledge distillation strategy and gives them state-of-the-art results. In contrast to their model, we focus on multi-modal reasoning and use reinforcement learning to improve the model performance beyond imitation learning, while utilizing all available data.

% \noindent \textbf{MARIO:} \citet{mario} release a new math dataset, enriched with the capability to use a Python code interpreter refined through a combination of GPT-4 annotations, human review, and self-training process.

%------------------------------------------------------------

\subsection{Code generating VLMs for general VQA}
\label{subsec:related_code_vlm}

While there have been approaches like ViperGPT \citep{vipergpt} and VizProg \citep{visual_programming} for generating code for tasks like visual question answering (VQA), they do so using LLMs, ignoring the image information completely and only looking at the question. Subsequently, they add the image information while executing the generated code. In contrast, our approach builds a VLM that looks at the geometry images and incorporates this information with the question to generate code for solving geometry problems.

% Prior works like ViperGPT \citep{vipergpt} and VizProg \citep{visual_programming} have used code generation for visual tasks such as visual question answering (VQA), but they rely solely on LLMs that process only the question text for generating code and later incorporate image information during code execution.
% % In contrast, our approach builds a VLM that integrates geometric image information alongside the question text to produce modular code for solving geometry problems.

% \citep{vipergpt, visual_programming}

% \noindent \textbf{Visual Programming:} \citet{visual_programming} propose VisProg which generates modular programs, which are then executed to get both the solution and the associated rationale.

% \noindent \textbf{ViperGPT:} \citet{vipergpt} present a framework that leverages code-generation LLMs to compose VLMs into subroutines which then produces a result the question. The model performs well in a zero-shot setting but does not involve any training.

% VisProg and ViperGPT are both similar to our step 1, described in section \ref{subsec:step_1}, but don't go beyond that and are specifically built for non-mathematical VQA.

%------------------------------------------------------------

\section{Proposed Method}
\label{sec:method}

%------------------------------------------------------------

\begin{figure*}[h]
\centering
\includegraphics[width=1\textwidth]{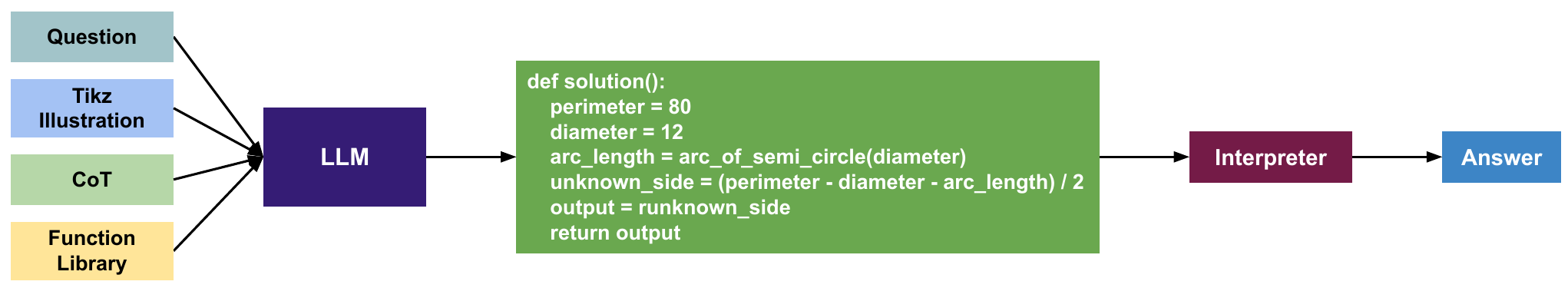}
\caption{The first step in our methodology consists of generating modular code by employing few-shot prompting with a code generation-capable LLM, utilizing questions, TikZ image illustrations, CoT reasoning, and the predefined function library. The generations that execute to produce the correct answer are selected as the basis for our "gold" code-tuning data, as discussed in \refsec{subsec:geo_data_gen}.}
\label{fig:step_1}
\end{figure*}

%%%%%%%%%%%%%%%%%%%%%%%%%%%%%%%%%%%%%

% In typical geometry problems, some elements (such as sides, angles, or areas) are provided as input, and the remaining elements are calculated step by step using geometric rules and formulas.
Existing approaches that finetune VLMs to generate CoT for geometry problems \citep{gllava,geomverse} help VLMs become better geometric reasoners, as compared to finetuning VLMs to generate just the final answer. %, as seen in \reftbl{tbl:results_geomverse}. 
 Even so, these CoT finetuned VLMs significantly suffer from wrong calculation and wrong formula usage errors, especially for geometry problems requiring long chains of reasoning \citep{geomverse}, due to the autoregressive nature of VLM language generation.
To address these limitations, we propose modular code-finetuning as an alternative to CoT finetuning for training VLMs to solve geometry problems.
The proposed approach, where the VLM generates Python code with calls to modules or functions from a predefined geometry library (discussed in \refsec{subsec:geo_data_gen}), offers three key advantages: (1) code execution guarantees precise and deterministic calculations%, in contrast to the stochastic nature of autoregressive next-token prediction
;
(2) the generated code references predefined mathematical functions, reducing errors related to incorrect formula usage; %by relying on a library that applies the correct formulas to compute results accurately
 and (3) using templated print statements in each function enhances the interpretability of the computational process.
We propose a two-step code-finetuning strategy, capitalizing on knowledge distillation, described in the following sections.%to solve geometric questions via modular code generation

%Our initial step involves generating "gold" code-tuning data by CoT few-shot prompting an LLM skilled in code generation, using the Tikz image illustrations \citep{tikz}, as explained in section \ref{subsec:geo_data_gen}.
%Subsequently, in section \ref{subsec:geo_model} we describe how we finetune a much smaller VLM on this LLM-generated "gold" code-tuning data, effectively teaching our VLM to generate modular code for solving complex geometry problems while looking at the image and the question. We call the resulting VLM - \model.
%In contrast to preceding models that generate code using an LLM and add the image information subsequently \citep{vipergpt, visual_programming}, our methodology is the first one to finetune a VLM specifically to generate code while looking at the image.

%In addition to \model, we propose \rag\ which is a retrieval-augmented version of \model\ based on the popular retrieval-augmented generation (RAG) paradigm \citep{rag_og}, as discussed in \refsec{subsec:geo_rag}. 

%%%%%%%%%%%%%%%%%%%%%%%%%%%%%%%%%%%%%

%%%%%%%%%%%%%%%%%%%%%%%%%%%%%%%%%%%%%

\subsection{Generating geometric code-tuning data}
\label{subsec:geo_data_gen}
%\paragraph{Geometry Functions Library}
%\label{subsec:geo_func}
% Our initial step involves generating 
We first generate "gold" code-tuning data by CoT few-shot prompting an LLM skilled in code generation, using the TikZ image illustrations \citep{tikz}.
% As a proof of concept, we manually construct a Python library comprising functions based on widely used geometry formulas. Furthermore, we translate the geometry formulas provided in the GeomVerse dataset into corresponding Python functions.
As illustrated in figure \ref{fig:step_1}, we start by prompting a large LLM to generate multiple Python codes that leverage pre-defined functions to solve geometry questions in a few-shot manner (6-shot in this study). We call this model our teacher LLM. 
We employ CoT prompting to guide the model generation, along with the TikZ illustration of the image \citep{tikz} (which is a textual representation of a geometry figure, as seen in the example in \reffig{fig:tikz_ill}) and a function library. We interpret the code generated by the model and use the answers it gets correct as a source of code-finetuning data for our student VLM described in section \ref{subsec:geo_model}. 
We also developed a Python function library that includes all formulas present in the GeomVerse dataset, as well as a selection of commonly applied geometry formulas.
Along with code for calculating answers using geometry formulas, the functions in the proposed library also include templated print statements. %that make the computations interpretable
%, discussed in \refsec{subsec:interpretability}.
The final library contains 47 geometry functions, which are listed in the Appendix. %This library can be further expanded with additional geometry functions or integrated with existing libraries, but we leave that for future work.
% This "gold" code training data is a subset of the full dataset as it only includes correct predictions from the "teacher" LLM.

%%%%%%%%%%%%%%%%%%%%%%%%%%%%%%%%%%%%%

%------------------------------------------------------------

\begin{figure*}[h]
\centering
\includegraphics[width=1\textwidth]{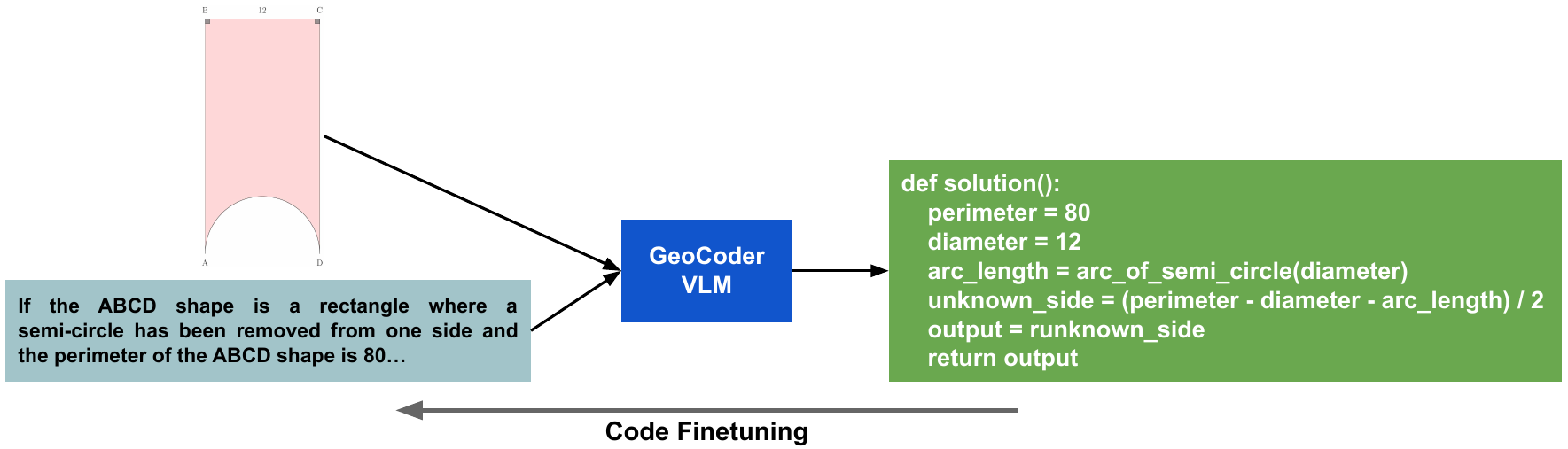}
\caption{During modular code-finetuning, we utilize the code-tuning data produced by our teacher LLM (see \refsec{subsec:geo_data_gen}) to finetune a significantly smaller VLM, which we refer to as \model\ (as discussed in \refsec{subsec:geo_model}).}
\label{fig:step_2}
\end{figure*}

%------------------------------------------------------------

\subsection{\model}
\label{subsec:geo_model}
% In the next step, 
%we finetune a much smaller VLM on this LLM-generated "gold" code-tuning data, effectively teaching our VLM to generate modular code for solving complex geometry problems while looking at the image and the question. We call the resulting VLM - \model.
We use the modular code-finetuning data generated by our teacher LLM (discussed in \refsec{subsec:geo_data_gen}) to finetune a much smaller VLM called the student VLM. As shown in figure \ref{fig:step_2}, the student VLM is finetuned to generate code directly given the image and textual question. We call the resulting code-finetuned student VLM, \model. 
% use LLaVA-1.5 7B \citep{llava} as the student VLM in our experiments and call the resulting code-tuned model, \model.
At inference time, \model\ takes the image and question as input and generates Python code which is interpreted to give the final answer.
In contrast to preceding models that generate code using an LLM and add the image information subsequently \citep{vipergpt, visual_programming}, our methodology is the first one to finetune a VLM specifically to generate code while looking at the image.
% This VLM is thus finetuned to generate code directly while looking at the image.
% The model learns to perform multi-hop reasoning by passing the output of one modular function as the input to another.

%------------------------------------------------------------

\begin{figure*}[h]
\centering
\includegraphics[width=1\textwidth]{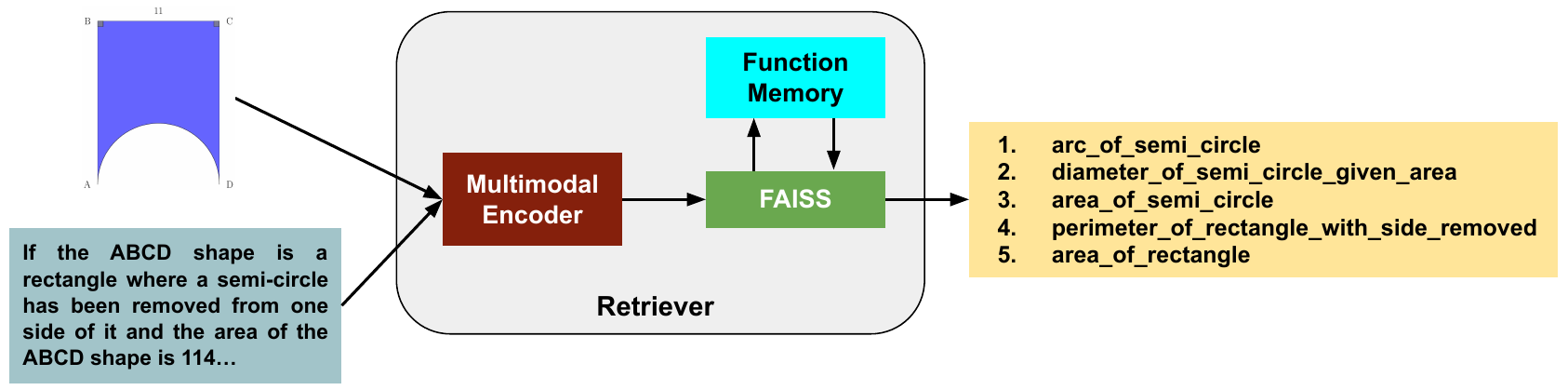}
\caption{For each geometry problem, given the image and question text, our multimodal retriever retrieves the most similar functions from the function memory, as discussed in \refsec{subsec:retriever}.}
\label{fig:rag_retriever}
\end{figure*}

%------------------------------------------------------------

\subsection{\rag}
\label{subsec:geo_rag}
% In addition to \model
Additionally, we propose \rag\ which is a retrieval-augmented version of \model\ based on the popular retrieval-augmented generation (RAG) paradigm \citep{rag_og}. \rag\ first employs a multimodal retriever (discussed in \refsec{subsec:retriever}) to retrieve relevant functions from the function library, which are then provided as input, along with the question and image, to the student VLM during both code-finetuning and inference.
This approach allows the student VLM to use a non-parametric memory to select the most appropriate functions, rather than relying on its parametric memory, thus increasing function usage (see \refsec{subsec:funtion_use}). 
% This provides two key benefits: (1) it enables the use of larger function libraries that would be impractical for the VLM to store in its memory, and (2) it allows \rag\ to adapt to updates in the function library without requiring further finetuning.

\subsubsection{Multimodal Retriever}
\label{subsec:retriever}
Given an image and accompanying question text from a geometry problem, our multimodal retriever identifies the most similar functions from the function memory.
As illustrated in \reffig{fig:rag_retriever}, the multimodal encoder first encodes the image and text into an embedding, and then FAISS \citep{faiss_lib,faiss_gpu} is used to retrieve a set of functions from a non-parametric memory whose embedding is closest to the input embedding.

\noindent \textbf{Multimodal Encoder.} In this work, we utilize the VISTA model \citep{vista} as our multimodal encoder, as it embeds text, image, and image-text data into a shared vector space, facilitating dense retrieval across different modalities.

\noindent \textbf{Function Memory.} We employ a non-parametric function memory structured as a dictionary, where multimodal encoder-generated embeddings of function names, arguments, and descriptions serve as keys, with the corresponding function names as values. This module stores a dictionary with embeddings and names for all functions in our library. %(described in \refsec{subsec:geo_func}).

% \noindent \textbf{Dense Retriever.} Given an image and accompanying question text from a geometry problem, our dense retriever identifies the most similar functions from the function memory.
% As illustrated in \reffig{fig:rag_retriever}, the multimodal encoder first encodes the image and text into an embedding, and then FAISS \citep{faiss_lib,faiss_gpu} is used to retrieve a set of functions whose embedding is closest to the input embedding.

\subsubsection{Finetuning and inference}
\label{subsec:ft_rag}
% As seen in \reffig{fig:rag_finetuning}, 
\rag\ student VLM receives the image, question text and a set of retrieved functions (details in \refsec{subsec:mmr}) and learns to generate modular code using the code-finetuning data generated by our teacher LLM (described in \refsec{subsec:geo_data_gen}). 
At inference time, we give the image, question and retrieved functions to \rag, which generates modular code which is executed to get the final answer.

\section{Experimental Setup}
\label{sec:setup}

%%%%%%%%%%%%%%%%%%%%%%%%%%%%%%%%%

%------------------------------------------------------------

\begin{table}[]
\centering
\begin{tabular}{|c|c|c|}
\hline
\textbf{Dataset} & \textbf{Train} & \textbf{Test} \\ \hline
GeomVerse & 10k & 3k \\ \hline
Geo170k-QA-NO & 110k & \xmark \\ \hline
GeoQA-NO & \xmark & 754 \\ \hline
\end{tabular}
\caption{Statistics of datasets used in this work, as discussed in \refsec{subsec:datasets}.}
\label{tbl:datasets}
\end{table}

%------------------------------------------------------------

\subsection{Datasets}
\label{subsec:datasets}
We evaluate our proposed models on two datasets: GeomVerse \citep{geomverse} and GeoQA \citep{geoqa}. Additionally, we use the Geo170k \citep{gllava} geometry instruction-tuning dataset and the train set of GeomVerse to create code-tuning data for finetuning \model, as described in \refsec{subsec:geo_data_gen}. 
An overview of the dataset statistics is outlined in \reftbl{tbl:datasets}.

% In GeoQA and Geo170k, each question is accompanied by four answer options for the model to select from, whereas in GeomVerse, the model must directly predict the final answer. Even though the final answer prediction task is much harder, we argue it is more realistic than selecting from options, as real-world geometry problems rarely provide answer choices. Therefore, we create and utilize 'no-option' versions of GeoQA and Geo170k in this study, where the task is to predict the final answer. 
% We call these versions GeoQA-NO and Geo170k-QA-NO.

% For every question in GeoQA and Geo170k, there exist four answer options for the model to choose from, while this is not the case in GeomVerse where the model has to predict the final answer directly. We believe the final answer prediction is a much more realistic task than option prediction as options are usually unavailable for most real-world geometry problems. Hence, we create and use 'no-option' versions of GeoQA and Geo170k in this work, where the task is the final answer prediction.

\subsubsection{Training Datasets}
\label{subsec:train_datasets}

\noindent \textbf{GeomVerse} 
is a synthetically generated dataset of geometry questions designed to require multi-step mathematical reasoning across both text and images. The dataset categorizes questions into three levels of complexity: depth 1, depth 2, and depth 3.
As seen in \reffig{fig:geomverse_depths}, depth 1 questions address a single geometric shape, while higher-depth questions involve multiple shapes, requiring a more complex reasoning process with multiple formula application steps. %, not additional factual knowledge.
Furthermore, all questions in GeomVerse are designed to necessitate models to look at the image to find the answer, as some critical information like relevant angles and side lengths are only mentioned in the image and not in the textual question.
The dataset contains $10k$ questions each in the train and validation sets. For our experiments on GeomVerse, we use the train set of GeomVerse to create a code-finetuning dataset, by generating 5 Python codes per geometry problem (using the process detailed in \refsec{subsec:geo_data_gen}), with $35k$ unique image-question and code pairs encompassing $8.5k$ unique geometry problems.

\noindent \textbf{Geo170k} is an instruction-tuning dataset divided into two components: alignment and QA-tuning, with no test set. The alignment portion of the dataset contains $60k$ geometry image-caption pairs, while QA-tuning includes $110k$ geometry question-answering problems derived by augmenting Geometry3k \citep{geometry3k} and GeoQA+ \citep{geoqa_plus}.
In Geo170k, each question is accompanied by four answer options for the model to select from, whereas in GeomVerse, the model must directly predict the final answer. Even though the final answer prediction task is much harder, we argue it is more realistic than selecting from options, as real-world geometry problems rarely provide answer choices. Therefore, we create and utilize a 'no-option' version of QA-tuning in this study, where the task is to predict the final answer. 
We call this version Geo170k-QA-NO, which contains $110k$ CoT-annotated questions, to finetune our models.
% In this study, we use the 'no-option' version of the GeoQA+ subset of QA-tuning, which contains $110k$ CoT-annotated questions, to finetune our models for evaluation on GeoQA-NO.
% We refer to this as the Geo170k-QA-NO dataset.
As there are no TikZ illustrations available in Geo170k, we use our teacher LLM to generate single code solutions for problems in Geo170k-QA-NO given only the CoT, question and function library using the method mentioned in \refsec{subsec:geo_data_gen}.
This creates a code-finetuning dataset, with $48k$ unique image-question and code pairs.

\subsubsection{Evaluation Datasets}
\label{subsec:eval_datasets}

\noindent \textbf{GeomVerse:} 
% is a synthetically generated dataset of geometry questions designed to require multi-step mathematical reasoning across both text and images. The dataset categorizes questions into three levels of complexity: depth 1, depth 2, and depth 3.
% As seen in \reffig{fig:geomverse_depths}, depth 1 questions address a single geometric shape, while higher-depth questions involve multiple shapes, requiring a more complex reasoning process with multiple formula application steps. %, not additional factual knowledge.
% Furthermore, all questions in GeomVerse are designed to necessitate models to look at the image to find the answer, as some critical information like relevant angles and side lengths are only mentioned in the image and not in the textual question. The 
We evaluate our models on the test set of GeomVerse which contains $1k$ questions from each depth, resulting in $3k$ test questions overall.

% is a dataset of synthetically generated geometry questions that require multi-hop mathematical reasoning over text and image. The questions are divided into three categories: depth 1, depth 2 and depth 3. Depth 1 problems contain questions about just one geometric shape, while additional shapes are added in higher-depth problems. Solving higher-depth problems requires long chains of reasoning rather than additional memorized knowledge. 

\noindent \textbf{GeoQA} includes geometry questions derived from Chinese middle school exams, with each problem annotated with the associated CoT (as seen in \reftbl{table:geo170k_cot_sample}).
The test set in GeoQA contains $754$ problems.
Since the original dataset is in Chinese, we follow \citet{gllava, unimath} and utilize the English version from \citet{geoqa_eng} to ensure consistency in language across datasets.
As with Geo170k, each question in GeoQA is provided with four answer options for the model to choose from.
In this work, we evaluate using GeoQA-NO, a version of the dataset where models predict the final answer without multiple-choice options.

%%%%%%%%%%%%%%%%%%%%%%%%%%%%%%%%%

\subsection{Metrics}
\label{subsec:metrics}
\noindent \textbf{Relaxed Accuracy:}
Following the methodology of \citet{geomverse}, \citet{chartqa_rel_acc} and \citet{plotqa_rel_acc}, we assess model performance based on relaxed accuracy, where a prediction is deemed correct if it is within three percent of the ground truth label.
This adjustment accounts for slight floating point computational differences arising from rounding during each calculation step.

%%%%%%%%%%%%%%%%%%%%%%%%%%%%%%%%%
\subsection{Our Models}
\label{subsec:mmr}
\noindent \textbf{Teacher LLM.} We use Llama 3.1 70B Instruct \citep{llama31} as our teacher LLM in this study because it is open-source and has strong code generation performance.

\noindent \textbf{Student VLM.} We use LLaVA 1.5 7B as the student VLM for code-finetuning \model\ and \rag\ in this study, as discussed in \refsec{sec:method}.
We generate 4 codes using beam search and select the first one that executes as the model's answer in our experiments. % with \model\ and \rag.

\noindent \textbf{Multimodal Retrieval.}
During training, we provide the VLM with k (here k= $7$) retrieved functions, in addition to the "gold" functions extracted from the corresponding "gold" code in the fine-tuning dataset. These "gold" functions are supplied during training to encourage the model to consistently select functions from the input list.
During inference, \rag\ is supplied with $10$ retrieved functions, where recall is $76\%$ (see \reffig{fig:prec_rec}). These choices are intended to optimize the recall of relevant functions while keeping the context length of the VLM manageable.
% The precision-recall curves for our function retriever can be seen in \reffig{fig:prec_rec}.

\subsection{Baselines}
\label{subsec:baselines}
In this study, we specifically compare the advantages of code-finetuning over CoT-finetuning.
On the GeomVerse dataset, we report results from experiments with three popular VLMs namely LLaVA 1.5 \citep{llava}, PaLI \citep{pali} and GPT4V \citep{gpt4}\footnote{The results on GPT4V are from \citet{geomverse} and were obtained on a subset of randomly selected 10 examples per depth, and the correctness was determined manually.}.
We consider five settings: zero-shot prompting (GPT4V), few-shot (4-shot) prompting with CoT (PaLI 55B), finetuning to predict the label directly (PaLI 5B), finetuning to predict CoT (PaLI 5B, which is the state-of-the-art on GeomVerse \citep{geomverse} and LLaVA 1.5 7B) and our proposed code-finetuning (\model\ 7B, \rag\ 7B), described in \refsec{sec:method}.
To show the difficulty of the GeomVerse dataset, we also report human accuracy on a small subset of the test split of the dataset from \citet{geomverse}.
In our experiments on GeoQA-NO, we compare code-finetuning with CoT-finetuning using the LLaVA 1.5 7B VLM.

%%%%%%%%%%%%%%%%%%%%%%%%%%%%%%%%%

\section{Results and discussion}
\label{sec:results}

%%%%%%%%%%%%%%%%%%%%%%%%%%%%%%%%%%%%%

%------------------------------------------------------------

\begin{table*}[]
\centering
\begin{tabular}{|c|c|c|c|c|}
\hline
\textbf{Model} & \textbf{Type}  & \textbf{Depth 1} & \textbf{Depth 2} & \textbf{Depth 3} \\ \hline
Human          & Human          & 80\%             & 65\%             & 55\%             \\ \hline
PaLI 55B       & CoT few-shot   & 0.1\%            & 0.3\%            & 0.2\%            \\
GPT4V          & CoT few-shot   & 70\%             & 0\%              & 0\%              \\ \hline
PaLI 5B        & Finetuned      & 22.8\%           & 15.6\%           & 14.5\%           \\
PaLI 5B        & CoT-finetuned  & 69.5\%           & 46.8\%           & 25.8\%           \\
LLaVA 1.5 7B   & CoT-finetuned  & 82\%             & 64\%             & 42\%             \\ \hline \hline
\model\ 7B (ours)    & Code-finetuned & 95.0\%           & 77.5\%           & \textbf{60.1\%}  \\
\rag\ 7B (ours) & Code-finetuned & \textbf{95.7\%}  & \textbf{80.1\%}  & 58.1\%           \\ \hline
\end{tabular}
\caption{Relaxed accuracy (described in \refsec{subsec:metrics}) on the GeomVerse dataset shows that our best-performing code-finetuned model, \rag, surpasses the corresponding CoT-finetuned LLaVA 1.5 by a margin of $13.7\%$ on depth 1 and $16.1\%$ on depth 2, while \model\ outperforms CoT-finetuned LLaVA 1.5 by $18.1\%$ on depth 3, as discussed in \refsec{subsec:quant}.}
\label{tbl:results_geomverse}
\end{table*}

%------------------------------------------------------------

\subsection{Quantitative Results}
\label{subsec:quant}

\noindent \textbf{Results on GeomVerse dataset:}
As seen in \reftbl{tbl:results_geomverse}, code-finetuning demonstrates significantly superior performance to CoT finetuned baselines on GeomVerse across depths, with or without retrieval-based augmentation.
\rag, outperforms CoT-finetuned PaLI 5B (the state-of-the-art on this dataset) by $26.2\%$ on depth 1 and $33.3\%$ on depth 2, while \model\ does so by $36.3\%$ on depth 3.
Code-finetuned \rag\ exceeds the performance of the corresponding CoT-finetuned LLaVA 1.5 by $13.7\%$ on depth 1 and $16.1\%$ on depth 2, while \model\ outperforms it by $18.1\%$ on depth 3.
%The key benefit of \rag\ over \model\ is its ability to handle larger libraries and adapt to library updates, yet it also demonstrates superior performance across all depth levels. 
The most significant improvement occurs at depths 2 and 3, which shows that modular code-finetuning enables language models to handle problems with long chains of mathematical reasoning and formula usage steps better than other finetuning approaches.

% with an advantage of nearly $4\%$ compared to \model.
% This further highlights the benefit of utilizing a non-parametric memory for storing function names, as opposed to depending on the model's parametric memory.
% It is interesting to note that both \model\ and \rag\ exceed human performance by a significant margin on depth 1 and depth 2 type problems.

%%%%%%%%%%%%%%%%%%%%%%%%%%%%%%%%%%%%%

%------------------------------------------------------------
\begin{table}[]
\centering
\begin{tabular}{|c|c|c|}
\hline
\textbf{Model} & \textbf{Type} & \textbf{\begin{tabular}[c]{@{}c@{}}Relaxed\\ Accuracy\end{tabular}} \\ \hline
LLaVA 1.5 7B & CoT-finetuned & 28.0\% \\ \hline
\model\ 7B & Code-finetuned & 42.3\% \\ \hline
\end{tabular}
\caption{Code-finetuned \model\ outperforms the corresponding CoT-finetuned LLaVA 1.5 by $14.3\%$ on the GeoQA-NO dataset, as discussed in \refsec{subsec:quant}.}
\label{tbl:geoqa_results}
\end{table}

%------------------------------------------------------------

\noindent \textbf{Results on GeoQA-NO dataset:}
Unlike the GeoQA dataset which has four answer choices for every geometry problem, in GeoQA-NO there are no options available and the task is to directly predict the final answer (as discussed in \refsec{subsec:datasets}).
As seen in \reffig{tbl:geoqa_results}, LLaVA 1.5 7B CoT finetuned on Geo170k-QA-NO (described in \refsec{subsec:datasets}), gets $28\%$ relaxed accuracy on the test set of GeoQA-NO.
% We think this low performance is because the CoTs provided in GeoQA-NO are not as detailed as the ones provided in the GeomVerse dataset (please refer to Appendix \ref{subsec:cotc_appendix} for examples).
In contrast, \model\ gets $42.3\%$ after code-finetuning on the gold code-tuning data made (discussed in \refsec{subsec:geo_data_gen}) using Geo170k-QA-NO.
Although this performance far exceeds (by $14.3\%$) that of the CoT-finetuned model, it remains relatively low. This indicates that GeoQA-NO is more challenging than GeoQA, as models tend to underperform when the task shifts from predicting options to predicting final answers.
Additionally, in contrast to GeomVerse, where the figures consist of combinations of relatively simple shapes and the CoTs are detailed (see \reftbl{table:geomverse_cot_sample}), GeoQA-NO and Geo170k-QA-NO involve figures that require more sophisticated image understanding. Additionally, the CoTs are brief and often lack clear explanations of the solution (see \reftbl{table:geo170k_cot_sample}).
This also results in open code and low function usage in the solutions produced by our teacher LLM. Hence we do not evaluate \rag\ in our experiments with GeoQA-NO.

% While figures in the GeomVerse dataset are composed of combinations are relatively simple shapes and the CoTs are detailed \reftbl{table:geomverse_cot_sample}, this is not the case in GeoQA-NO where the figures require more complex image understanding and the CoTs are short and don't always explain the solution well (as seen in \reftbl{table:geo170k_cot_sample}).
% Additionally, we think this low performance of models on GeoQA-NO as compared to their performance on GeomVerse can be attributed to the 
% Additionally, we think this low performance is also because the CoTs provided in GeoQA-NO (used for CoT finetuning, as well as to generate code-finetuning data) are not as detailed as the ones provided in the GeomVerse dataset (please refer to Appendix \ref{subsec:cotc_appendix} for examples).

%%%%%%%%%%%%%%%%%%%%%%%%%%%%%%%%%%%%%

\subsection{Error Analysis}
\label{subsec:error}

%------------------------------------------------------------

\begin{table*}[]
\centering
\begin{tabular}{|ccccccccc|}
\hline
\multicolumn{1}{|c|}{} & \multicolumn{1}{c|}{\begin{tabular}[c]{@{}c@{}}Correct\\ Output\end{tabular}} & \multicolumn{1}{c|}{\begin{tabular}[c]{@{}c@{}}Wrong\\ Output\end{tabular}} & \multicolumn{1}{c|}{\begin{tabular}[c]{@{}c@{}}Error\\ Count\end{tabular}} & \multicolumn{1}{c|}{\begin{tabular}[c]{@{}c@{}}Math-domain\\ Errors\end{tabular}} & \multicolumn{1}{c|}{\begin{tabular}[c]{@{}c@{}}Name\\ Errors\end{tabular}} & \multicolumn{1}{c|}{\begin{tabular}[c]{@{}c@{}}Syntax\\ Errors\end{tabular}} & \multicolumn{1}{c|}{\begin{tabular}[c]{@{}c@{}}Zero Div.\\ Errors\end{tabular}} & \begin{tabular}[c]{@{}c@{}}Other\\ Errors\end{tabular} \\ \hline
\multicolumn{9}{|c|}{\model} \\ \hline
\multicolumn{1}{|c|}{Depth 1} & 950 & \multicolumn{1}{c|}{38} & \multicolumn{1}{c|}{12} & 0 & 2 & 9 & 1 & 0 \\ \cline{1-1}
\multicolumn{1}{|c|}{Depth 2} & 775 & \multicolumn{1}{c|}{211} & \multicolumn{1}{c|}{14} & 0 & 0 & 13 & 0 & 1 \\ \cline{1-1}
\multicolumn{1}{|c|}{Depth 3} & 601 & \multicolumn{1}{c|}{365} & \multicolumn{1}{c|}{34} & 2 & 9 & 21 & 0 & 2 \\ \hline
\multicolumn{9}{|c|}{\rag} \\ \hline
\multicolumn{1}{|c|}{Depth 1} & 957 & \multicolumn{1}{c|}{43} & \multicolumn{1}{c|}{0} & 0 & 0 & 0 & 0 & 0 \\ \cline{1-1}
\multicolumn{1}{|c|}{Depth 2} & 801 & \multicolumn{1}{c|}{178} & \multicolumn{1}{c|}{21} & 0 & 17 & 2 & 0 & 2 \\ \cline{1-1}
\multicolumn{1}{|c|}{Depth 3} & 581 & \multicolumn{1}{c|}{369} & \multicolumn{1}{c|}{50} & 3 & 34 & 2 & 1 & 10 \\ \hline
\end{tabular}
\caption{Error analysis of \model\ and \rag\ on the test set of GeomVerse with $1000$ examples per depth, as discussed in \refsec{subsec:error}.}
\label{tbl:error_analysis}
\end{table*}

%------------------------------------------------------------

In this study, 'wrong outputs' refer to cases where the generated code runs but yields an incorrect answer, whereas 'error count' refers to instances where the code fails to execute and throws an exception. These exceptions are classified into five categories: math-domain errors (input values outside the function’s domain), name errors (calling a function not present in the library), syntax errors (incorrect Python syntax), zero-division errors, and other uncategorized exceptions.
As seen in \reftbl{tbl:error_analysis}, on the GeomVerse dataset, \rag\ reduces the amount of syntax errors compared to \model, but the number of name errors increases.
% \rag\ reduces wrong outputs compared to \model\ by 46 and 23 instances for depth 2 and 3, respectively, across 1000 problems per depth.
% While \rag\ eliminates exceptions entirely for depth 1 problems, it introduces more exceptions in depth 2 and 3 problems compared to \model. 
\rag\ introduces more exceptions in depth 2 and 3 problems compared to \model. 
The primary cause of this increase is the rise in name errors for higher-depth problems. We attribute this to the relatively low recall rate of approximately $76\%$ during multimodal retrieval at test time (see \reffig{fig:prec_rec}).
In depth 2 and 3 problems, where multiple functions are needed within a single solution, missing functions in the prompt lead the model to select incorrect ones, unlike \model\ which relies on its parametric memory.

%%%%%%%%%%%%%%%%%%%%%%%%%%%%%%%%%%%%%

%------------------------------------------------------------

\begin{figure*}[h]
\centering
\includegraphics[width=1\textwidth]{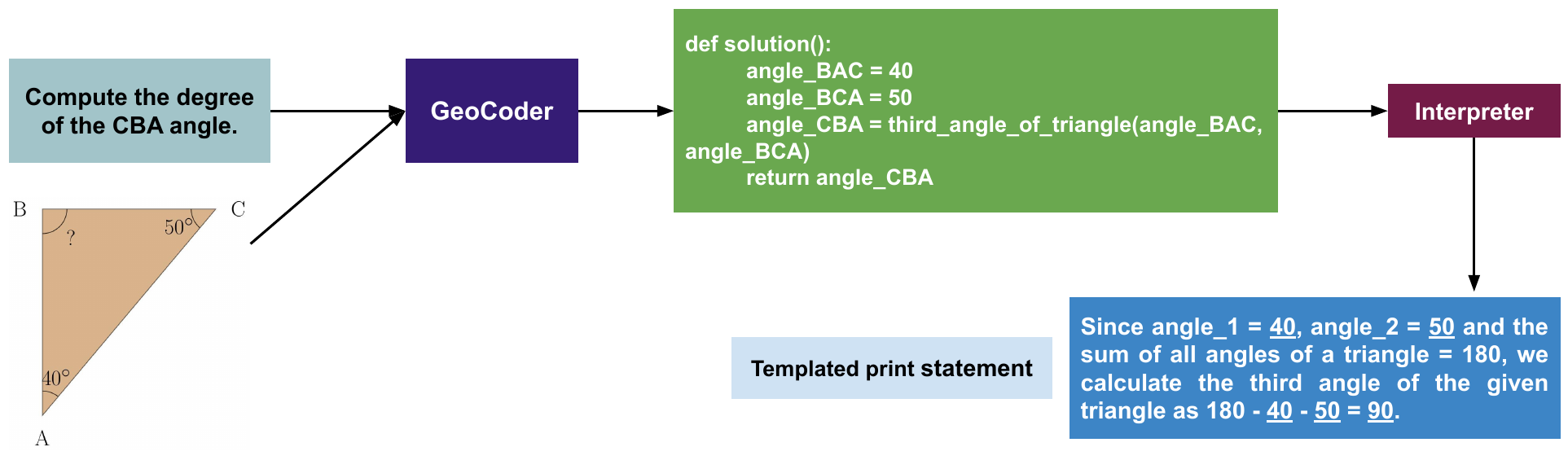}
\caption{Modular functions add interpretability, as discussed in \refsec{subsec:interpretability}. In this example, the underlined values are filled in the template by the '$third\_angle\_of\_triangle$' function.}
\label{fig:interpretability}
\end{figure*}

%------------------------------------------------------------

\subsection{Interpretability}
\label{subsec:interpretability}
In addition to geometric formula calculations, the functions in our proposed function library % (discussed in \refsec{subsec:geo_func}) 
 integrate templated print statements, making the computation process more interpretable.
As seen in the example in \reffig{fig:interpretability}, the manually created templates in our functions are filled in automatically with the values from the arguments of the function and the computed results, thus introducing deterministic interpretability.
This contrasts with the stochastic interpretability provided by CoT solutions where the autoregressive language model can hallucinate the rationale behind the computation and still arrive at the correct result.

%------------------------------------------------------------

\begin{table}[]
\centering
\begin{tabular}{|c|c|c|c|}
\hline
\textbf{Model} & \textbf{D1} & \textbf{D2} & \textbf{D3} \\ \hline
\model\ & 71.7\% & 72.6\% & 70.6\% \\ \hline
\rag\ & 90.3\% & 85.2\% & 90.0\% \\ \hline
\end{tabular}
\caption{Function usage analysis. \rag\ uses functions for $17\%$ more questions on average across depths as compared to \model\ on the test set of GeomVerse, as discussed in \refsec{subsec:funtion_use}.}
\label{tbl:function_use_analysis}
\end{table}

%------------------------------------------------------------

\subsection{Function Usage Analysis}
\label{subsec:funtion_use}
Functions add interpretability (see \refsec{subsec:interpretability}) and ensure correct geometry formula usage.
To understand how well our models use functions, we compare the number of times they use functions to solve a given geometry problem.
Specifically, we find that out of 1000 questions per depth in the GeomVerse test set, \rag\ uses functions to answer $903$ problems in depth 1, $852$ in depth 2 and $900$ in depth 3.
In contrast, \model\ uses functions for only $717$ problems in depth 1, $726$ in depth 2 and $706$ in depth 3.
As seen in \reftbl{tbl:function_use_analysis}, \rag\ demonstrates a $17\%$ higher function usage on average across different depths, compared to \model\ on the GeomVerse dataset.
This shows the effectiveness of \rag\ at generating modular, interpretable code for solving geometry problems.

% % their performance on the subset of questions from GeomVerse where they use functions.
% Specifically, we find that out of 1000 questions per depth in the GeomVerse test set, \rag\ uses functions to answer $845$ problems in depth 1, $722$ in depth 2 and $103$ in depth 3.
% We report the performance of our models on this subset of problems in \reftbl{tbl:function_use_analysis}.
% From the table, we see that when using functions, \rag\ outperforms \model\ by $46.7\%$ on depth 1 and $22\%$ on depth 2, while the performance on depth 3 remains the same.
% This shows the effectiveness of \rag\ on using functions to solve geometry problems.

%%%%%%%%%%%%%%%%%%%%%%%%%%%%%%%%%%%%%
\section{Conclusion}
\label{sec:conclusion}

In this work, we present \model, a novel modular framework for code-finetuning aimed at solving geometry problems using VLMs.
Our experiments demonstrate that code-finetuning outperforms CoT-finetuning for training autoregressive VLMs, as the deterministic calculations provided by code execution overcome the stochastic nature of VLM-based next-token prediction.
To apply correct geometry formulas, mitigate errors from incorrect usage, and increase interpretability, \model\ uses a predefined function library that encodes widely used geometry formulas.
In addition, we present \rag, a retrieval-augmented variant that uses a non-parametric memory module to retrieve functions from the geometry function library, avoiding dependence on the model’s parametric memory.
% \rag\ offers the advantage of handling larger libraries and adapting to changes, while also surpassing \model\ in performance.
The code-finetuned models we propose consistently outperform CoT-finetuned models across question complexity and different datasets.
\section{Limitations}
\label{sec:limitations}

Although \model\ is designed to utilize functions from our predefined geometry library, it is not strictly required to do so while solving problems, as there is no enforced constraint on function usage. In the prompt provided to the teacher LLM, we encourage the use of functions from the library, but the model is not restricted to generating code solely from the listed functions.
While the majority of the "gold" code produced by the teacher LLM includes library functions, some instances deviate from this, marking a limitation in our method.
As a result, on the GeomVerse test set, \rag\ does not use any functions for $9.7\%$ problems in depth 1, $14.8\%$ problems in depth 2 and $10\%$ problems in depth 3 (see \reftbl{tbl:function_use_analysis}).
Another limitation of our work is that the proposed geometric function library may not cover all the formulas required for solving every geometry question. However, our approach is designed to accommodate larger libraries and external function APIs to remedy this. We leave that exploration for future work.
Moreover, since the vision modules in VLMs are typically pre-trained on real-world imagery, they struggle with interpreting geometric figures. The images in GeomVerse and GeoQA-NO differ significantly, requiring us to finetune our models on their respective training datasets in this study. 
This limitation may be addressed as future VLMs are pre-trained on a more diverse set of geometric figures.

\section*{Acknowledgements}
\label{sec:acknowledgements}
We thank NSERC and Samsung for supporting this work and CIFAR for their support under the Canada CIFAR AI Chair program. We would also like to thank Aishwarya Agrawal and the class of IFT 6765 for their valuable feedback during the early development of this work.

\bibliography{GeoCoder/references}

\appendix

\section{Appendix}
\label{sec:appendix}
%You may include other additional sections here.

\subsection{Geometry Functions Library}
\label{subsec:geo_func_appendix}

Here is the list of functions in our proposed function library along with their descriptions:

% \begin{itemize}
\begin{enumerate}

  \item radius\_of\_sector\_given\_arc\_length(arc\_length, angle) \# Calculates the radius of a sector given the angle and the arc length.
  \item side\_of\_parallelogram\_given\_area(area, side, angle) \# Calculates the second side of a parallelogram given the area, the first side and the sine of the angle (in degrees) between them.
  \item height\_of\_triangle\_given\_area(area, base) \# Calculates the height of a triangle given the area and the base.
  \item base\_of\_triangle\_given\_area(area, height) \# Calculates the base of a triangle given the area and the height.
  \item arc\_length\_of\_sector(angle, radius) \# Calculates arc length of a sector given the angle and the radius.
  \item length\_of\_third\_side(side\_1, side\_2, angle) \# Calculates the length of the third side of a triangle given two sides and the angle between them in degrees, according to the law of cosines.
  \item height\_of\_equilateral\_triangle(base) \# Calculates the height of an equilateral triangle given the base, according to the sine rule.
  \item diagonal\_of\_square(side) \# Calculates diagonal of a square given side.
  \item hypotenuse\_of\_right\_triangle(side\_1, side\_2) \# Calculates hypotenuse of a right triangle given two sides, according to Pythagorean theorem.
  \item side\_of\_right\_triangle\_given\_side(side, hypotenuse) \# Calculates the second side of a right triangle given the first side and hypotenuse, according to Pythagorean theorem.
  \item diameter\_of\_semi\_circle\_given\_perimeter (perimeter) \# Calculates the diameter of a semi-circle given the perimeter/circumference.
  \item side\_of\_right\_triangle\_given\_angle (opposite\_angle, hypotenuse) \# Calculates the length of the side of a right triangle given opposite angle (in degrees) and hypotenuse.
  \item side\_of\_equilateral\_triangle(height) \# Calculates the side of an equilateral triangle given the height, according to the sine rule.
  \item side\_of\_square\_given\_area(area) \# Calculates the side of a square given the area.
  \item arc\_of\_semi\_circle(diameter) \# Calculates the arc length or arc circumference of a semi-circle given the diameter.
  \item diagonal\_of\_rectangle(side\_1, side\_2) \# Calculates the diagonal of a rectangle given two sides, according to Pythagorean theorem.
  \item side\_of\_rectangle\_given\_diagonal(side, diagonal) \# Calculates the second side of a rectangle given the first side and diagonal, according to the Pythagorean theorem.
  \item diameter\_of\_semi\_circle\_given\_area(area) \# Calculates the diameter of a semi-circle given the area.
  \item area\_of\_rectangle(side\_1, side\_2) \# Calculates area of a rectangle given two sides.
  \item area\_of\_square\_given\_side(side) \# Calculates the area of a square given a side.
  \item area\_of\_square\_given\_diagonal(diagonal) \# Calculates the area of a square given the diagonal.
  \item area\_of\_triangle\_given\_base\_height(base, height) \# Calculates the area of a triangle given the base and height.
  \item area\_of\_right\_triangle(side\_1, side\_2) \# Calculates the area of a right triangle given two perpendicular sides.
  \item area\_of\_semi\_circle(diameter) \# Calculates the area of a semi-circle given the diameter.
  \item area\_of\_circle(diameter) \# Calculates the area of a circle given the diameter.
  \item area\_of\_parallelogram(side\_1, side\_2, angle) \# Calculates the area of a parallelogram given two sides and the sine of the angle (in degrees) between them.
  \item area\_of\_triangle\_given\_semi\_perimeter (semi\_perimeter, side\_1, side\_2, side\_3) \# Calculates the area of a triangle given three sides and the semi-perimeter, using Heron's formula.
  \item area\_of\_trapezoid(base\_1, base\_2, height) \# Calculates the area of a trapezoid given the two bases and the height.
  \item area\_of\_equilateral\_triangle(side) \# Calculates the area of an equilateral triangle given the side.
  \item area\_of\_sector(angle, radius) \# Calculates the area of a sector given the angle and radius.
  \item side\_of\_square\_given\_perimeter(perimeter) \# Calculates the length of a side of a square given the perimeter.
  \item perimeter\_of\_shape(sides) \# Calculates the perimeter of a shape given a list of sides.
  \item perimeter\_of\_square(side) \# Calculates the perimeter of a square given a side.
  \item perimeter\_of\_rectangle\_with\_side\_removed
  (side\_1, side\_2) \# Calculates perimeter of a rectangle with one side (side\_2) removed given two sides.
  \item perimeter\_of\_square\_given\_diagonal(diagonal) \# Calculates the perimeter of a square given the diagonal.
  \item perimeter\_of\_trapezoid(base\_1, base\_2, side\_1, side\_2) \# Calculates the perimeter of a trapezoid given both the bases and the lateral sides.
  \item perimeter\_of\_rectangle(width, height) \# Calculates the perimeter of a rectangle given both the width and the height.
  \item perimeter\_of\_parallelogram(side\_1, side\_2) \# Calculates perimeter of a parallelogram given two sides.
  \item perimeter\_of\_triangle(side\_1, side\_2, side\_3) \# Calculates perimeter of a triangle given all three sides.
  \item perimeter\_of\_semi\_circle(diameter) \# Calculates the perimeter/circumference of a semi-circle given the diameter.
  \item third\_angle\_of\_triangle(angle\_1, angle\_2) \# Calculates the third of a triangle given the other two angles.
  \item tangent\_of\_angle(angle) \# Calculates the tangent of an angle. The tangent of an angle in a right triangle gives the ratio between the opposite and the adjacent sides.
  \item angle\_of\_right\_triangle(opposite\_side, hypotenuse) \# Calculates the angle (in degrees) of a right triangle given the opposite side and hypotenuse.
  \item complementary\_angle(angle) \# Calculates the complementary angle for a given angle.
  \item angle\_of\_sector(arc\_length, radius) \# Calculates the angle of a sector given the radius and the arc length.
  \item ngle\_of\_parallelogram\_given\_area(side\_1, side\_2, area) \# Calculates angle between two sides of a parallelogram given two sides and the area.
  \item solve\_for\_x(eq\_1, eq\_2) \# Solves a system of equations to get value of x.

\end{enumerate}
% \end{itemize}

%------------------------------------------------------------

\begin{figure*}[h]
\centering
\includegraphics[width=0.6\textwidth]{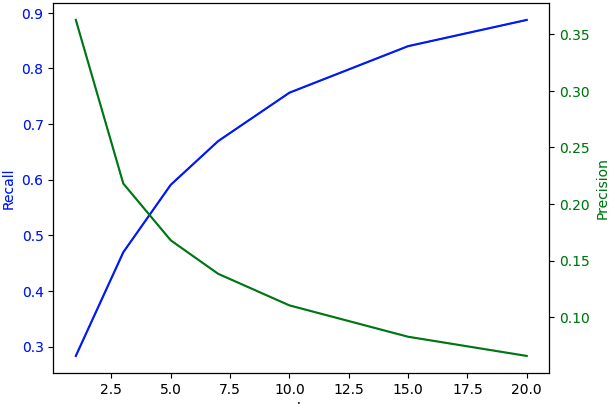}
\caption{Precision-Recall for multimodal retrieval in \rag\ on GeomVerse dataset.}
\label{fig:prec_rec}
\end{figure*}

%------------------------------------------------------------

%------------------------------------------------------------

\begin{figure*}[h]
\centering
\includegraphics[width=1\textwidth]{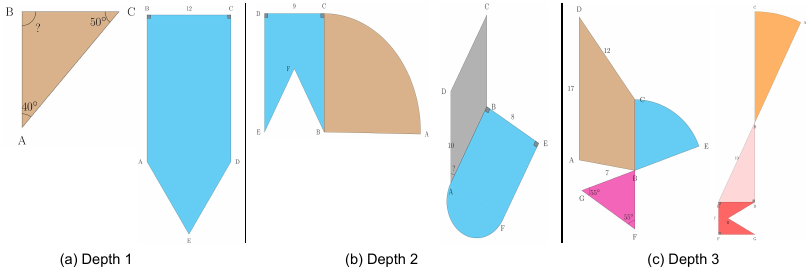}
\caption{Complexity levels at different depths in the GeomVerse dataset.}
\label{fig:geomverse_depths}
\end{figure*}

%------------------------------------------------------------
\subsection{Hyper Parameters Used}
% In this study, we use hyperparameters to optimize the performance of the student model.
In this work, we used 4-bit QLoRA \citep{qlora} to finetune all our models. The LoRA Rank and LoRA Alpha \citep{lora} values were set to 128 and 256, respectively, to control the rank of the low-rank adaptation layers, allowing efficient fine-tuning. A Projection Learning Rate of 2e-5 was chosen to ensure stable gradient updates during training. The model was trained for 10 epochs, with an initial learning rate of 2e-4, following a cosine decay schedule that progressively reduced the learning rate for smoother convergence. A learning rate warmup ratio of 0.03 was employed to avoid sharp transitions at the start of training, preventing instability. Additionally, no weight decay was applied, as regularization through this method was unnecessary for the dataset and the size of the model used.

\subsection{Additional Figures}
\label{subsec:tikz_appendix}

\begin{figure*}[h!]
\centering
\includegraphics[width=0.75\textwidth]{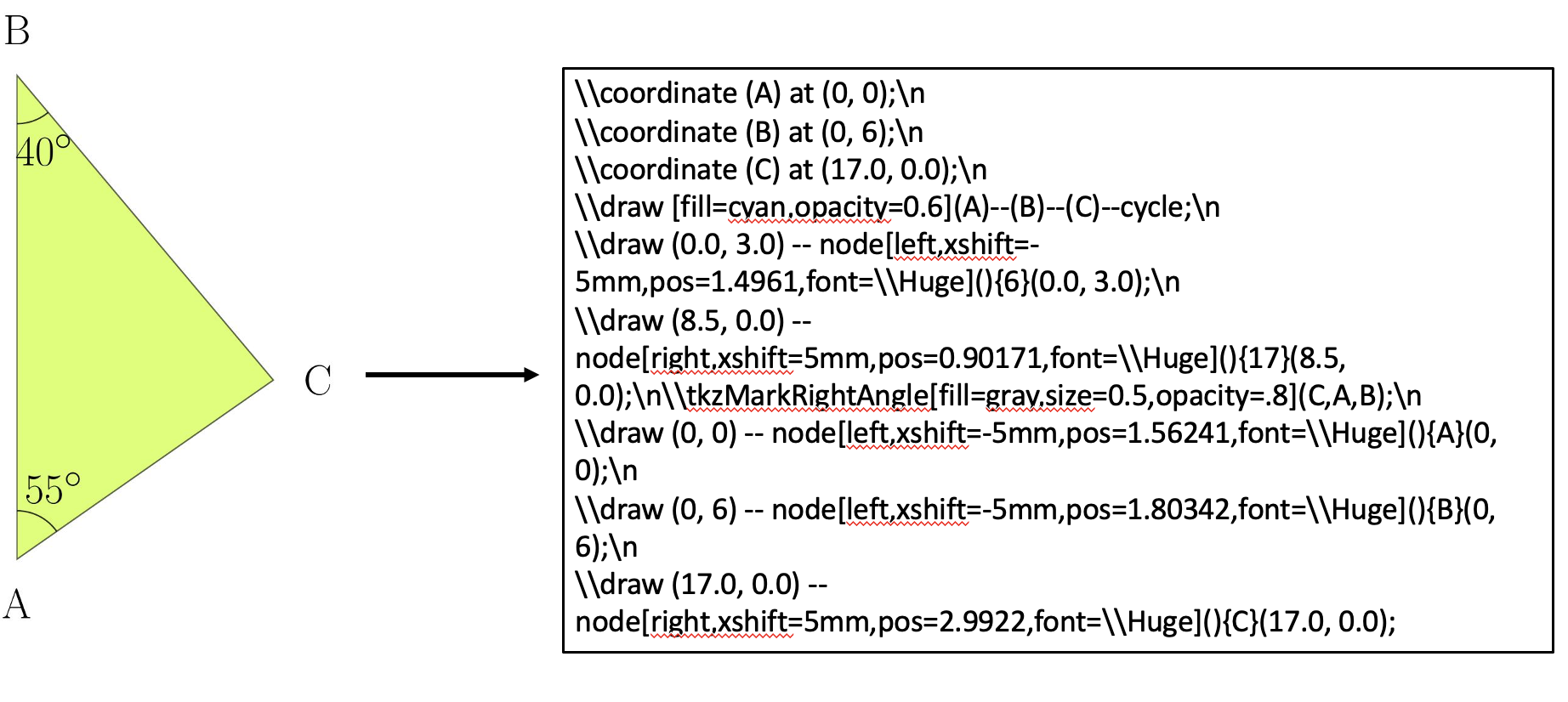}
\caption{TikZ representation of an image from the GeomVerse dataset.}
\label{fig:tikz_ill}
\end{figure*}

% \subsection{Chain-of-Thought examples from the GeomVerse and GeoQA-NO dataset}
% \label{subsec:cotc_appendix}

\begin{table*}[h!]
    \centering
    \begin{tabular}{m{3.5cm}|m{11cm}}  
        \toprule
        \multicolumn{1}{c|}{Image} & \multicolumn{1}{c}{Chain-of-Thought} \\ [0.5ex]
        \midrule
        \includegraphics[width=0.15\textwidth]{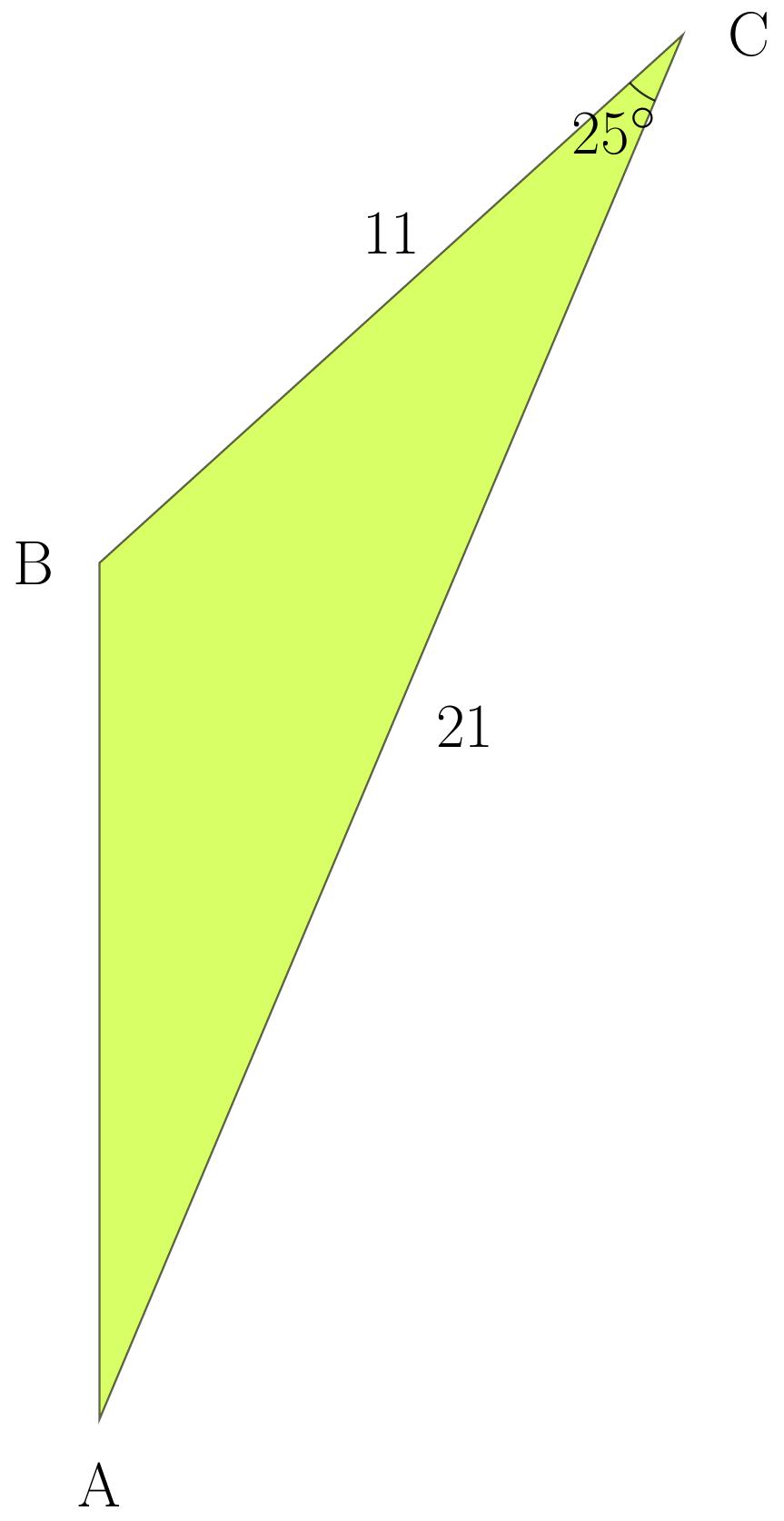} & 
        \ttfamily  % Use monospaced font
        \textbf{Question:}
        Compute the length of the AB side of the ABC triangle. Round computations to 2 decimal places.
        
        \textbf{Chain-of-Thought:}
        For the ABC triangle, the lengths of the AC and BC sides are 21 and 11 and the degree of the angle between them is 25. Therefore, the length of the AB side is equal to \$\textbackslash \textbackslash sqrt{21\^{}2 + 11\^{}2 - (2 * 21 * 11) * \textbackslash \textbackslash cos(25)} = \textbackslash \textbackslash sqrt{441 + 121 - 462 * (0.91)} =  \textbackslash \textbackslash sqrt{562 - (420.42)} = \textbackslash \textbackslash sqrt{141.58} = 11.9\$. Therefore the final answer is 11.9.
        \\
        \midrule
        \includegraphics[width=0.23\textwidth]{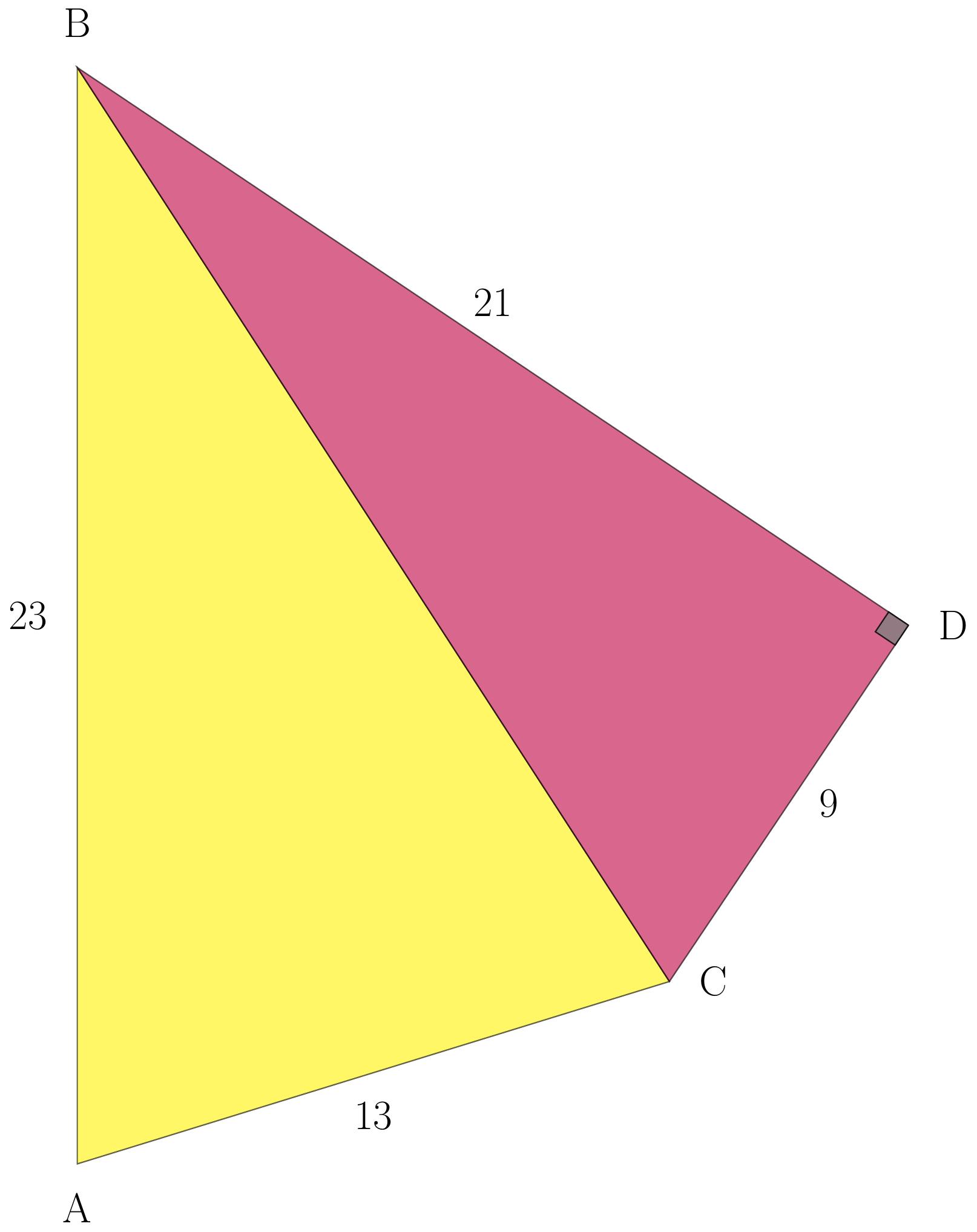} & 
        \ttfamily  % Use monospaced font
        \textbf{Question:}
        Compute the area of the ABC triangle. Round computations to 2 decimal places.
        
        \textbf{Chain-of-Thought:}
        The lengths of the BD and CD sides of the BCD triangle are 21 and 9, so the length of the hypotenuse (the BC side) is \$\textbackslash \textbackslash sqrt{21\^{}2 + 9\^{}2} = \textbackslash \textbackslash sqrt{441 + 81} = \textbackslash \textbackslash sqrt{522} = 22.85\$. We know the lengths of the AB, AC and BC sides of the ABC triangle are 23 and 13 and 22.85, so the semi-perimeter equals \$(23 + 13 + 22.85) / 2 = 29.43\$. So the area is \$\textbackslash \textbackslash sqrt{29.43 * (29.43-23) * (29.43-13) * (29.43-22.85)} = \textbackslash \textbackslash sqrt{29.43 * 6.43 * 16.43 * 6.58} = \textbackslash \textbackslash sqrt{20458.07} = 143.03\$. Therefore the final answer is 143.03.
        \\
        \midrule
        \includegraphics[width=0.23\textwidth]{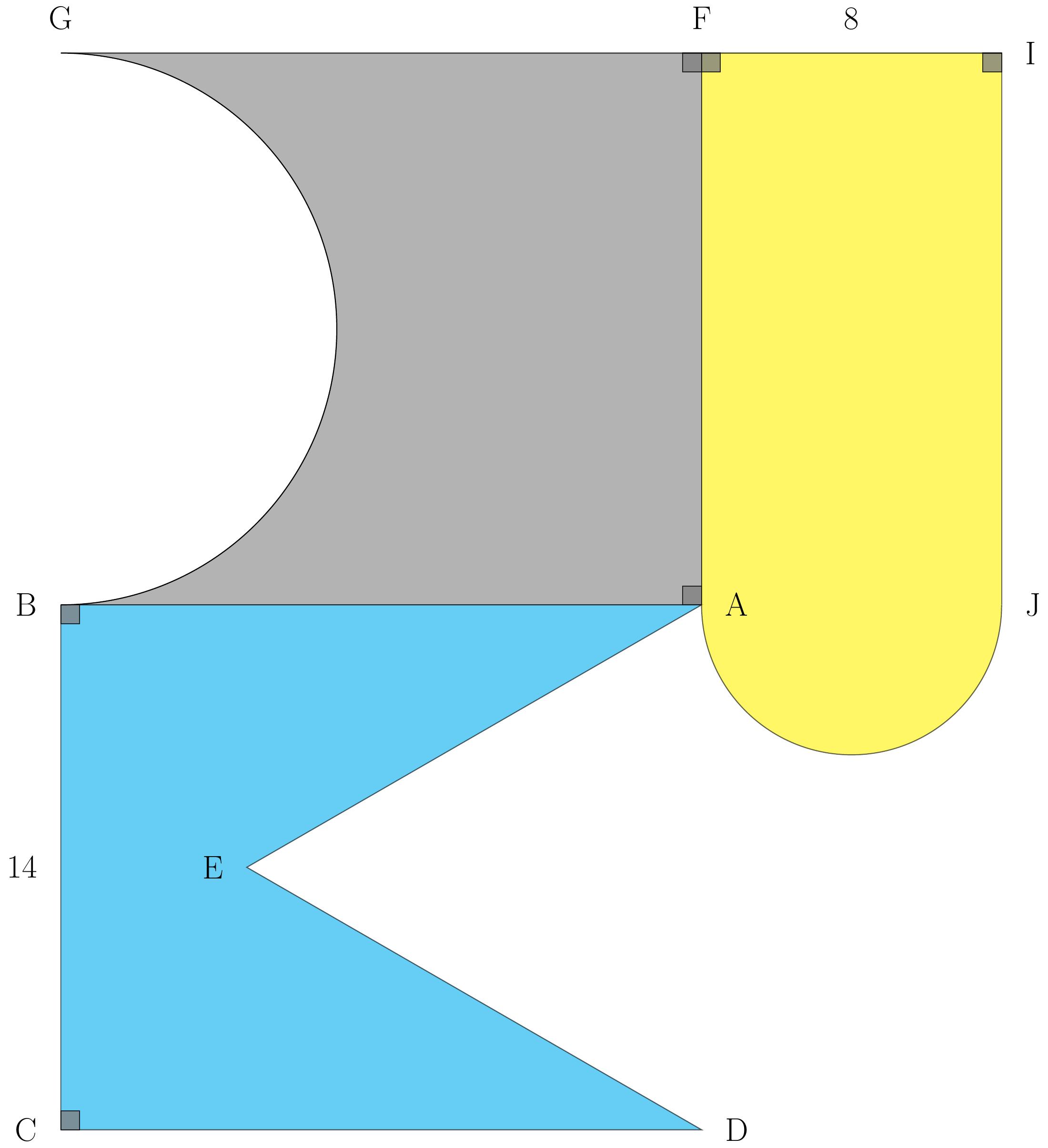} &
        \ttfamily  % Use monospaced font
        \textbf{Question:} 
        If the ABCDE shape is a rectangle where an equilateral triangle has been removed from one side of it, the BAFG shape is a rectangle where a semi-circle has been removed from one side of it, the perimeter of the BAFG shape is 72, the AFIJ shape is a combination of a rectangle and a semi-circle and the perimeter of the AFIJ shape is 50, compute the perimeter of the ABCDE shape. Assume $\pi$=3.14. Round computations to 2 decimal places.
        
        \textbf{Chain-of-Thought:}
        The perimeter of the AFIJ shape is 50 and the length of the FI side is 8, so \$2 * OtherSide + 8 + \textbackslash \textbackslash frac{8 * 3.14}{2} = 50\$. So \$2 * OtherSide = 50 - 8 - \textbackslash \textbackslash frac{8 * 3.14}{2} = 50 - 8 - \textbackslash \textbackslash frac{25.12}{2} = 50 - 8 - 12.56 = 29.44\$. Therefore, the length of the AF side is \$\textbackslash \textbackslash frac{29.44}{2} = 14.72\$. The diameter of the semi-circle in the BAFG shape is equal to the side of the rectangle with length 14.72 so the shape has two sides with equal but unknown lengths, one side with length 14.72, and one semi-circle arc with diameter 14.72. So the perimeter is \$2 * UnknownSide + 14.72 + \textbackslash \textbackslash frac{14.72 * \textbackslash \textbackslash $\pi$}{2}\$. So \$2 * UnknownSide + 14.72 + \textbackslash \textbackslash frac{14.72 * 3.14}{2} = 72\$. So \$2 * UnknownSide = 72 - 14.72 - \textbackslash \textbackslash frac{14.72 * 3.14}{2} = 72 - 14.72 - \textbackslash \textbackslash frac{46.22}{2} = 72 - 14.72 - 23.11 = 34.17\$. Therefore, the length of the AB side is \$\textbackslash \textbackslash frac{34.17}{2} = 17.09\$. The side of the equilateral triangle in the ABCDE shape is equal to the side of the rectangle with width 14 so the shape has two rectangle sides with length 17.09, one rectangle side with length 14, and two triangle sides with lengths 14 so its perimeter becomes \$2 * 17.09 + 3 * 14 = 34.18 + 42 = 76.18\$. Therefore the final answer is 76.18
        \\
        \bottomrule
    \end{tabular}
    \caption{Sample Chain-of-Thought from the GeomVerse dataset}
    \label{table:geomverse_cot_sample}
\end{table*}

\begin{table*}[h!]
    \centering
    \begin{tabular}{m{3.5cm}|m{11cm}}  
        \toprule
        \multicolumn{1}{c|}{Image} & \multicolumn{1}{c}{Chain-of-Thought} \\ [0.5ex]
        \midrule
        \includegraphics[width=0.23\textwidth]{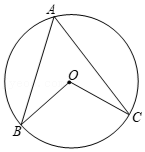} & 
        
        \ttfamily  % Use monospaced font
        \textbf{Question:} In the provided diagram, if points A, B, and C all lie on the circle O and angle BAC measures 54°, what is the measure of angle BOC? 
        
        \textbf{Chain-of-Thought:} since angle A = 54°, therefore, angle BOC = 2 angle A = 108°. Therefore, the answer is 108°
        \\
        \midrule
        \includegraphics[width=0.23\textwidth]{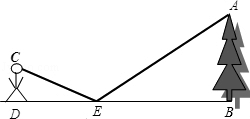} & 
        \ttfamily  % Use monospaced font
        \textbf{Question:} What method did the mathematics interest group use to measure the height of a tree on the horizontal ground of the campus as depicted in the figure?
        
        \textbf{Chain-of-Thought:} Explanation: Given that angle AEB=angle CED, we can conclude that Right Triangle ABE is similar to Right Triangle CDE (using the property of angles). Therefore, the ratio of their corresponding sides must be equal: AB/CD = BE/DE. Substituting the given values, we have AB/1.6 = 7.8/3.2. Solving for AB, we get AB = 3.9 (meters).
        \\
        \midrule
        \includegraphics[width=0.23\textwidth]{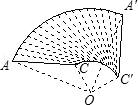} & 
        \ttfamily  % Use monospaced font
        \textbf{Question:} Given the configuration, let the length of the windshield wiper AC be denoted as `L`. If AO = 65.0, CO = 15.0, and the wiper AC rotates 90.0 around the point O, determine the expression for the area swept by the wiper AC in terms of `L` and $\pi$.
        
        \textbf{Chain-of-Thought:} The area swept by the wiper AC can be calculated using the concept of congruent triangles and sectors. Substituting the given values, the area can be represented as ((65\^{}2 - 15\^{}2)/4) $\pi$. By substituting `L` for 65, the area becomes ((L\^{}2 - 15\^{}2)/4) $\pi$, expressed in terms of `L`. Thus, the area swept by the wiper AC is 1000 $\pi$ cm\^{}2.
        \\
        \bottomrule
    \end{tabular}
    \caption{Sample Chain-of-Thought from the Geo170k-QA-NO and GeoQA-NO datasets.}
    \label{table:geo170k_cot_sample}
\end{table*}

\begin{table*}[h!]
    \centering
    \begin{tabular}{m{2.5cm}|m{12cm}}  
        \toprule
        \multicolumn{1}{c|}{Image} & \multicolumn{1}{c}{\rag\ Output} \\ [0.5ex]
        \midrule
        \includegraphics[width=0.15\textwidth]{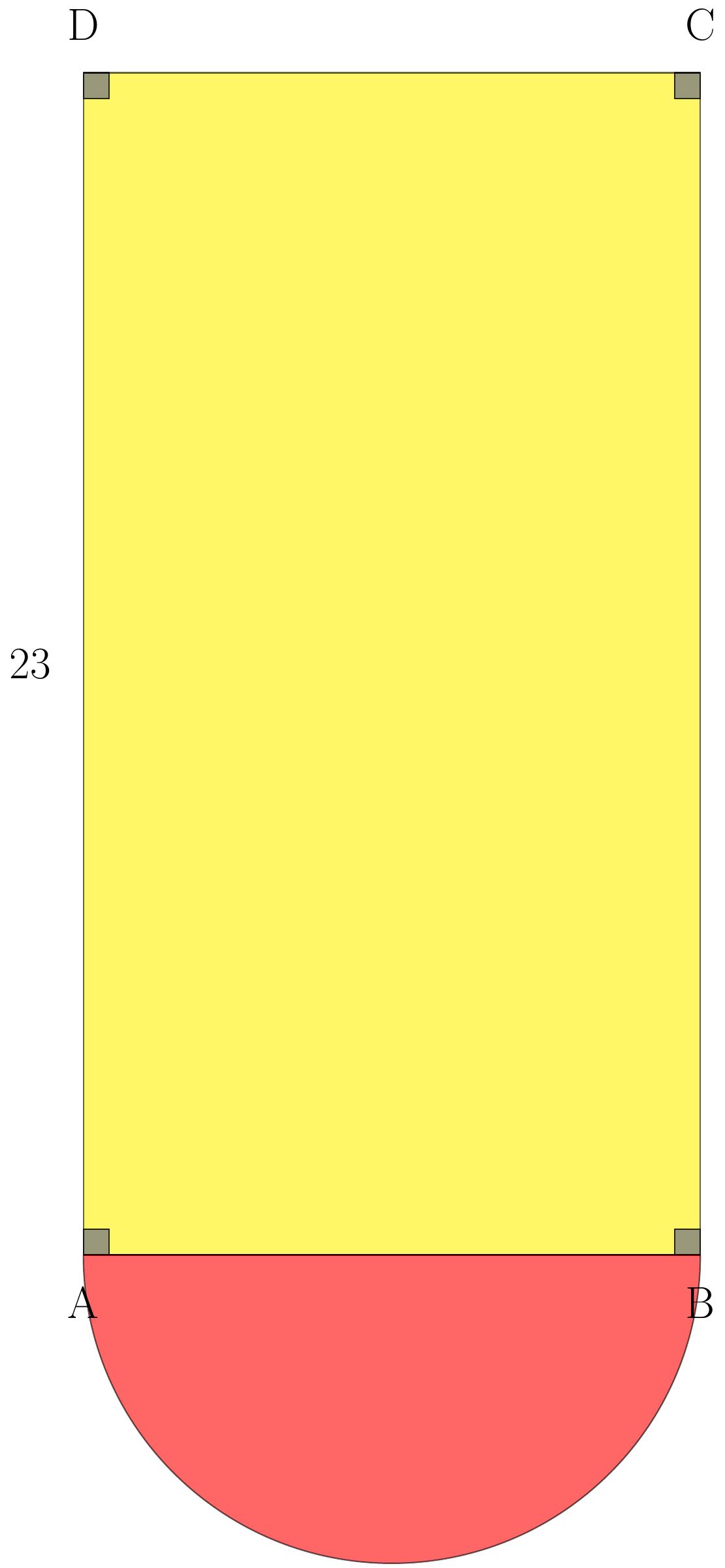} & 
        
        \ttfamily  % Use monospaced font
        {\small
        \textbf{Question:} If the area of the red semi-circle is 56.52, compute the diagonal of the ABCD rectangle. 

        \textbf{Ground Truth:} 25.94
        
        \textbf{RAG-GeoCoder:}
{\small
\begin{verbatim}
def solution():
    area_red_semicircle = 56.52
    diameter = diameter_of_semi_circle_given_area(area_red_semicircle)
    side_AD = 23
    diagonal = diagonal_of_rectangle(side_AD, diameter)
    return diagonal
\end{verbatim}
}
        \textbf{Code Output:} 25.94
        }
        \\
        \midrule
        \includegraphics[width=0.17\textwidth]{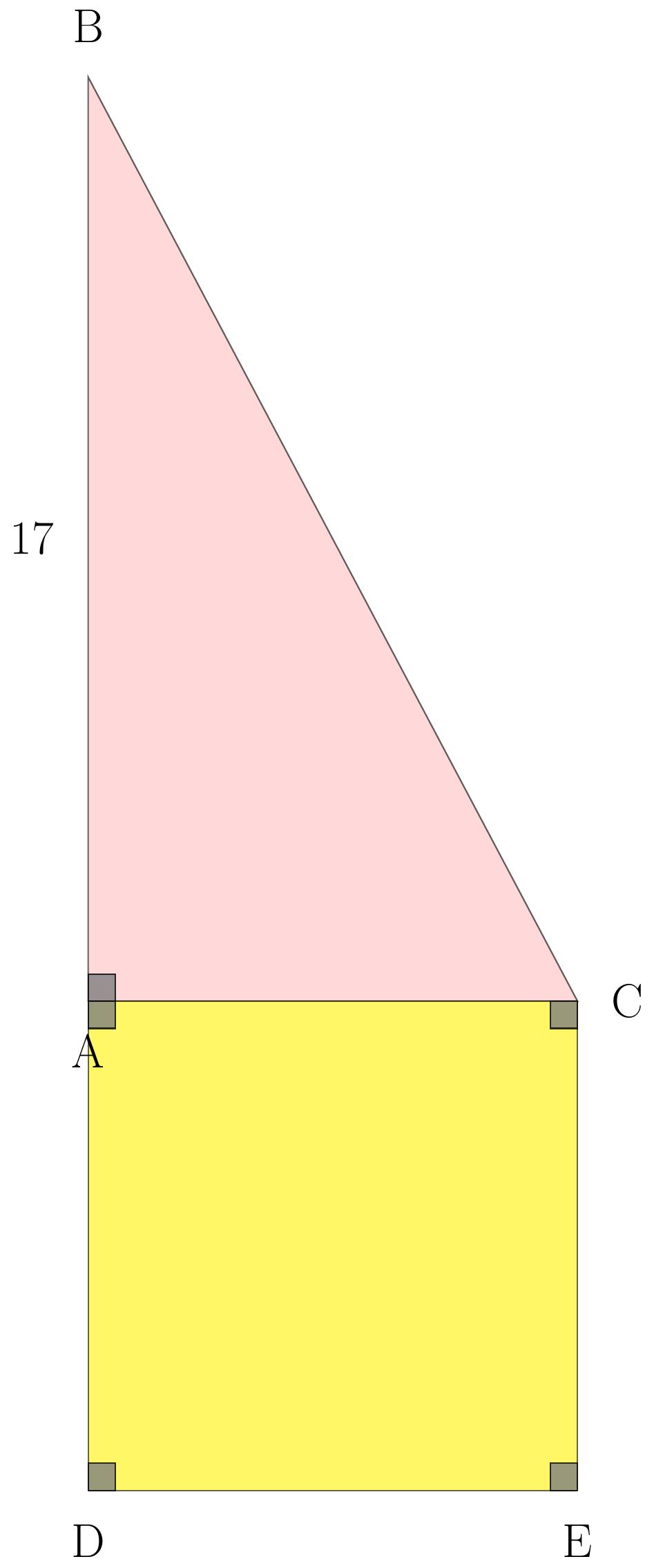} & 
        
        \ttfamily  % Use monospaced font
        {\small
        \textbf{Question:} If the area of the ADEC square is 81, compute the area of the ABC right triangle.

        \textbf{Ground Truth:} 76.5
        
        \textbf{RAG-GeoCoder:}
{\small
\begin{verbatim}
def solution():
    area_ADEC = 81
    side_ADEC = side_of_square_given_area(area_ADEC)
    side_AB = 17
    area_ABC = area_of_triangle_given_base_height(side_ADEC, side_AB)
    output = area_ABC
    return output
\end{verbatim}
}
        \textbf{Code Output:} 76.5
        }
        \\
        \midrule
        \includegraphics[width=0.17\textwidth]{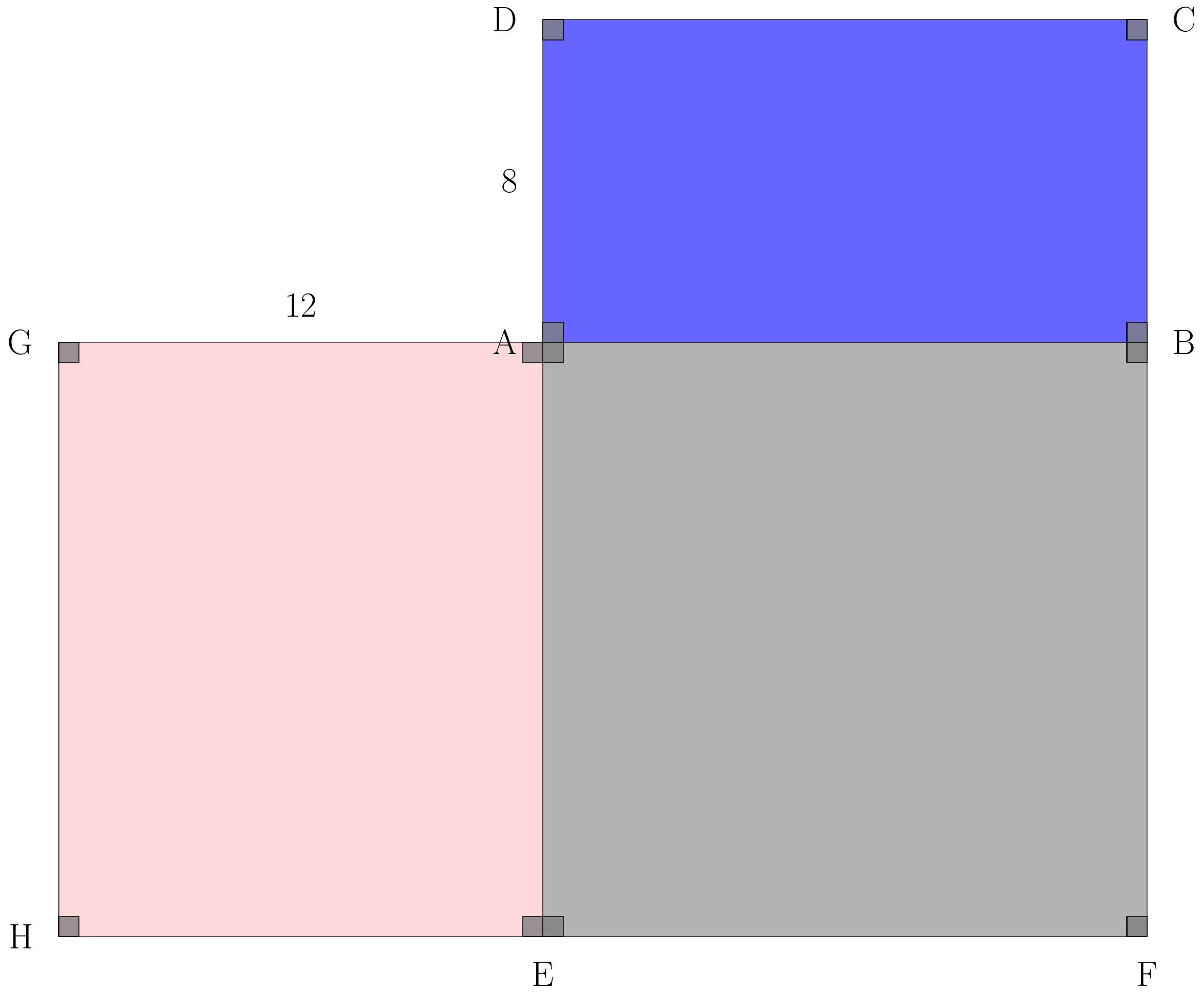} & 
        
        \ttfamily  % Use monospaced font
        {\small
        \textbf{Question:} If the diagonal of the AEFB rectangle is 21 and the diagonal of the AGHE rectangle is 19, compute the diagonal of the ABCD rectangle.

        \textbf{Ground Truth:} 16.97
        
        \textbf{RAG-GeoCoder:}
{\small
\begin{verbatim}
def solution():
    diagonal_AEFB = 21
    diagonal_AGHE = 19
    side_AG = 12
    side_AE = side_of_rectangle_given_diagonal(diagonal_AGHE, side_AG)
    side_AB = side_of_rectangle_given_diagonal(diagonal_AEFB, side_AE)
    side_AD = 8
    output = diagonal_of_rectangle(side_AB, side_AD)
    return output
\end{verbatim}
}
        \textbf{Code Output:} 16.97
        }
        \\
        \midrule
        \includegraphics[width=0.17\textwidth]{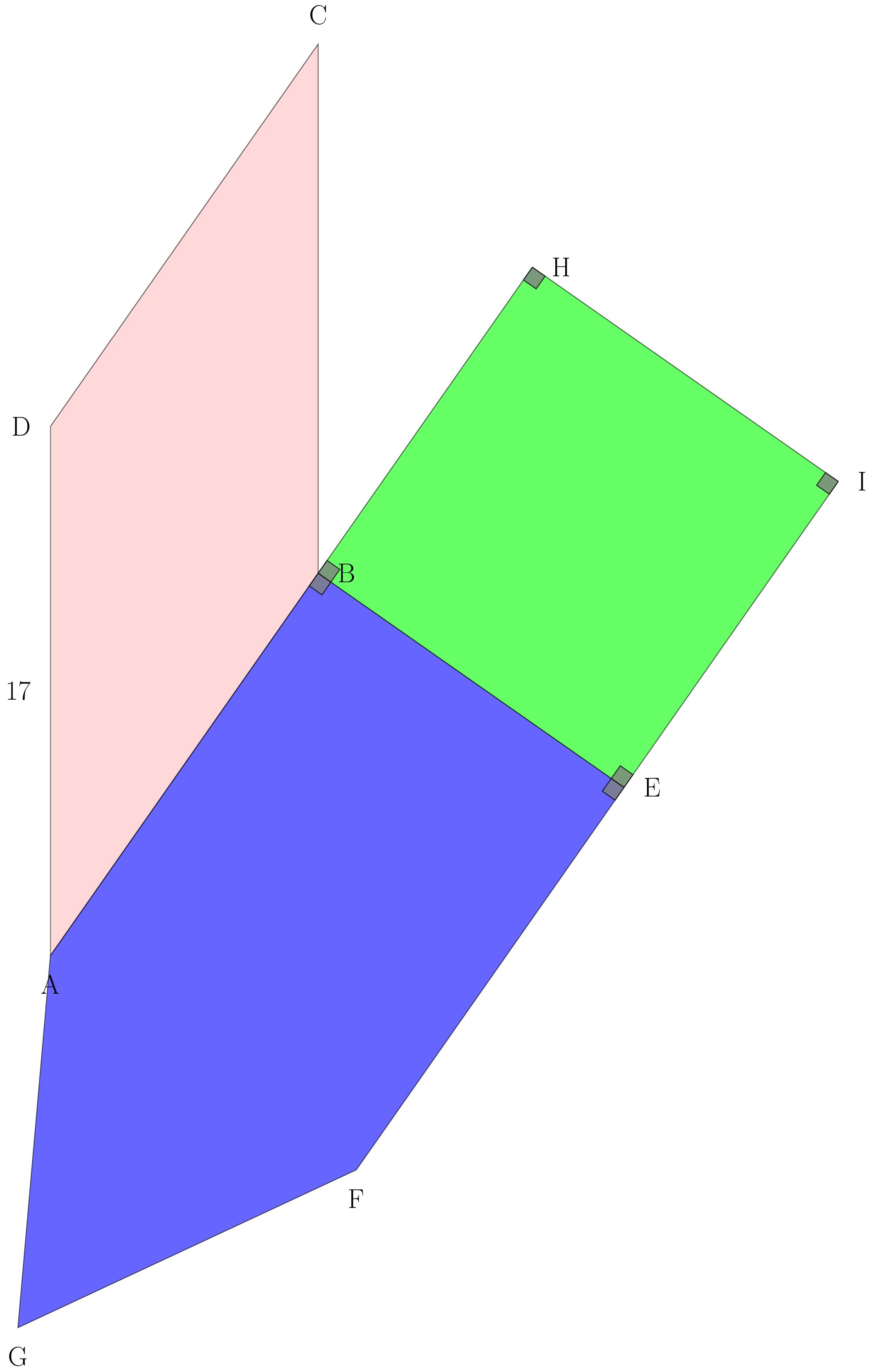} & 
        
        \ttfamily  % Use monospaced font
        {\small
        \textbf{Question:} If the ABEFG shape is a combination of a rectangle and an equilateral triangle, the perimeter of the ABEFG shape is 66 and the perimeter of the BHIE square is 48, compute the perimeter of the ABCD parallelogram.

        \textbf{Ground Truth:} 64.0
        
        \textbf{RAG-GeoCoder:}

\begin{verbatim}
def solution():
    perimeter_ABEFG = 66
    perimeter_BHIE = 48
    side_BE = side_of_square_given_perimeter(perimeter_BHIE)
    side_AB = (perimeter_ABEFG - 3 * side_BE) / 2
    side_AD = 17
    output = perimeter_of_parallelogram(side_AB, side_AD)
    return output
\end{verbatim}

        \textbf{Code Output:} 64.0
        }
        \\
        \bottomrule
    \end{tabular}
    \caption{Sample code generations by the \rag\ model on the GeomVerse test set.}
    \label{table:sample_geocoder_output}
\end{table*}

\end{document}